\documentclass[lettersize,journal]{IEEEtran}
\usepackage{amsmath,amsfonts}
\usepackage{algorithmic}
\usepackage{algorithm}
\usepackage{booktabs}
\usepackage{array}
\usepackage[caption=false,font=normalsize,labelfont=sf,textfont=sf]{subfig}
\usepackage{textcomp}
\usepackage{stfloats}
\usepackage{url}
\usepackage{verbatim}
\usepackage[table]{xcolor}
\usepackage{graphicx}
\usepackage{tikz}
\usepackage{cite}
\usepackage{float}
\usepackage{multirow}
\usepackage{caption}
\usepackage{tabularx}
\definecolor{oursblue}{RGB}{222,235,247}
\definecolor{gtgray}{RGB}{238,238,238}
\usepackage[pagebackref=false,breaklinks=true,colorlinks=true,bookmarks=false,citecolor=blue,linkcolor=red]{hyperref}
\usepackage{cleveref}

\crefname{section}{Sect.}{Sects.}
\crefname{figure}{Fig.}{Figs.}
\crefname{table}{Tab.}{Tabs.}

\begin{document}

\title{SocialHumanoid: Towards Expressive Humanoid Behavior via One-Step Co-Speech Motion Generation}

\author{Chengqun~Yang, Tengjie~Zhu, Liang~Xu, Fulong~Liu, Guanzhu~Ren, Yitong~Xing, Xuefeng~Lu, Fei~Shi, Siyuan~Fan, Weijie~Dong, Yao~Mu, Xiaokang~Yang,~\IEEEmembership{Fellow,~IEEE} and Yichao~Yan

\thanks{Chengqun~Yang, Tengjie~Zhu, Liang~Xu, Fulong~Liu, Guanzhu~Ren, Yitong~Xing, Yao~Mu, Xiaokang~Yang and Yichao~Yan are with Shanghai Jiao Tong University. (Email: \{ycq0191, zhutengjie, liangxu, lfl925884898, guanzhuren, richardxyt, muyao, xkyang, yanyichao\}@sjtu.edu.cn)}
\thanks{Xuefeng~Lu, Fei~Shi, Siyuan~Fan and Weijie~Dong are with ZTE Corporation. (Email: lu.xuefeng@zte.com.cn, Disk.shi@163.com, fsyxjtuecl@gmail.com, weijiedong1949@gmail.com)}
\thanks{Corresponding author: Yichao Yan.}
}

\maketitle
\pagestyle{plain}       
\thispagestyle{plain}

\begin{abstract}
Humanoid robots are increasingly expected to serve as embodied social agents that communicate naturally with humans through face-to-face interaction. 
During such communication, humanoid robots require body behaviors that are synchronized with speech, affectively expressive, and suitable for real-time execution. 
However, existing co-speech methods are primarily developed for digital humans and lack joint support for affective control and low-latency continuous generation on physical embodiments.
To bridge this gap, we present \textbf{SocialHumanoid}, a system for expressive humanoid behavior via one-step co-speech motion generation.
Given response speech and a specified affective condition, SocialHumanoid generates each full-body motion window in a single forward pass and connects successive windows through motion-history conditioning.
The generated human motion is further converted online into embodiment-compatible robot references and tracked by a whole-body controller for physical execution.
To provide explicit supervision for affective body expression, we further introduce \textbf{AffectMoCap}, a 4-hour dataset captured from two professional actors, containing synchronized speech, body motion, fine-grained hand motion, and emotion annotations.
On BEAT2, SocialHumanoid achieves the best FGD among the compared generation methods, competitive speech-motion synchrony, and approximately $\mathbf{6}\times$ faster inference than GestureLSM under the same protocol.
Perceptual evaluations further show that training with AffectMoCap improves affect recognition from generated body motion, while real-robot experiments demonstrate continuous affect-conditioned behavior and stable long-horizon execution.
Our project page is \url{https://rex0191.github.io/SocialHumanoid/}.
\end{abstract}


\section{Introduction}
\label{sec:intro}
\IEEEPARstart{W}{ith} the rapid advancement of embodied intelligence, humanoid robots are moving from task-oriented machines toward embodied agents that communicate naturally with people.
During face-to-face communication, hand gestures, torso movement, posture, and motion rhythm complement speech by conveying emphasis, intention, attitude, and emotion~\cite{yoon2020speech,li2021audio2gestures,beat,liu2022learning}.
A robot may therefore produce a coherent spoken response yet still appear rigid or disengaged when its body remains static, repetitive, or emotionally uniform.
For physical humanoids, expressive co-speech behavior should not only remain synchronized with speech and respond visibly to affective conditions, but also be generated continuously and remain suitable for robot execution.
\begin{figure}[t]
\centering
\includegraphics[width=\columnwidth]{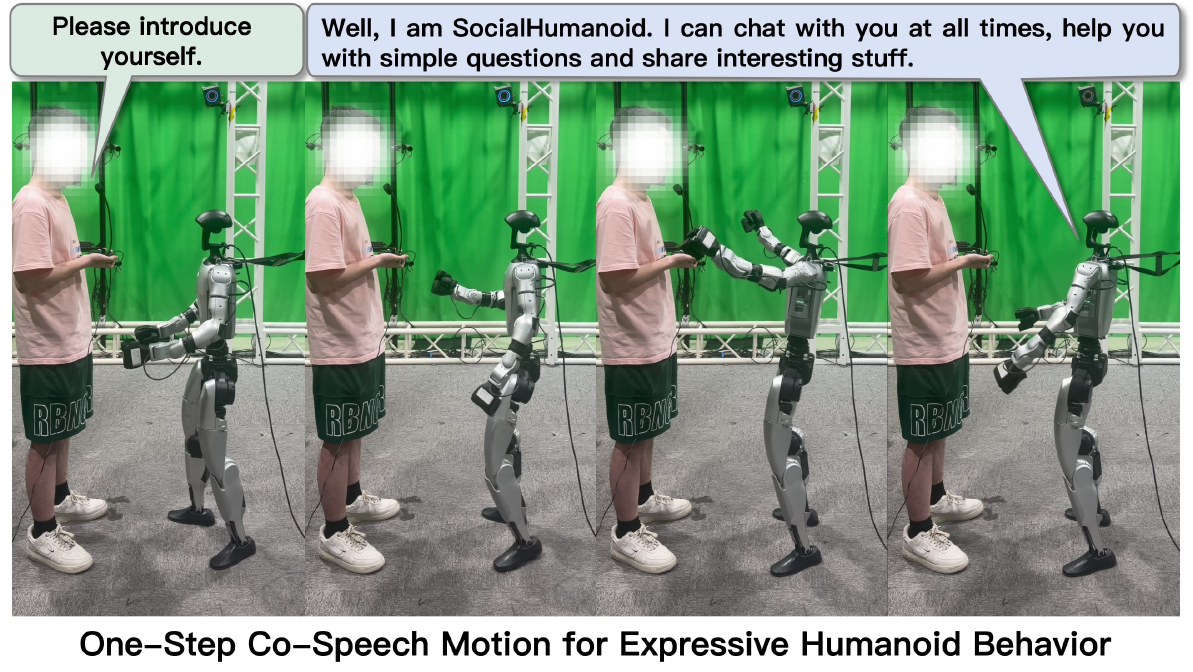}
\caption{\textbf{Expressive humanoid behavior during a robot response.} SocialHumanoid converts a spoken response and an affective condition into continuous full-body co-speech motion through one-step inference, then retargets and executes the motion online on a physical humanoid.}
\label{fig:teaser}
\vspace{-1.0em}
\end{figure}
Recent co-speech generation methods~\cite{zhu2023taming,ao2023gesturediffuclip,emage,diffsheg,gesturelsm,zhang2025semtalk,fang2026coordspeaker} have substantially improved motion naturalness, diversity, and alignment with speech rhythm or semantics.
Most, however, are developed for digital humans and evaluated primarily in human skeletal or parametric motion spaces.
Moreover, diffusion and iterative flow-based models may require multiple inference steps for each motion window, which is undesirable when motion references must be generated continuously during robot speech.
Recent systems improve efficiency through shortcut, autoregressive, or streaming formulations~\cite{gesturelsm,miburi,livegesture,proact}, but low-latency affect-conditioned motion generation suitable for humanoid robots remains relatively under-explored.
Physical deployment further requires the generated human motion to be adapted to the robot embodiment and tracked through whole-body control~\cite{araujo2025retargeting,ji2024exbody2,luo2025sonic,pan2025spider,huang2026omg}.
\IEEEpubidadjcol
Affective expressiveness presents another challenge due to limited supervision.
Plausible speech-aligned gestures are not necessarily affectively distinguishable, and learning controllable body expression requires consistent examples of how emotion affects gesture amplitude, motion energy, posture, and hand articulation.
BEAT~\cite{beat} provides emotion annotations and ZeroEGGS~\cite{ghorbani2023zeroeggs} provides diverse style labels, but professional performances combining explicit emotion conditions with synchronized speech, full-body motion, and detailed hand motion remain limited.

To address these challenges, we present \textbf{SocialHumanoid}, a system for expressive humanoid behavior via one-step co-speech motion generation.
Given response audio and a specified affective condition, SocialHumanoid generates continuous full-body SMPL-X motion in a part-wise residual vector-quantized latent space.
We adapt improved MeanFlow~\cite{geng2026improved} to emotion-conditioned co-speech synthesis, enabling one-step inference for each latent motion window.
A short history from the preceding window is reused as a seed, allowing successive windows to form a continuous motion stream.
The generated motion is then retargeted online into embodiment-compatible robot references and executed through whole-body control on a humanoid robot.
After each synchronized robot response, the dialogue and speech frontend can receive the next user utterance and repeat the same response pipeline, allowing the system to operate across multiple dialogue turns.

To strengthen affective supervision, we further collect \textbf{AffectMoCap}, a dataset containing 4 hours of synchronized speech, full-body motion, fine-grained hand motion, and eight emotion labels from two professional actors.
The dataset captures professional performances under explicit emotion conditions to provide consistent supervision for variations in motion energy, gesture amplitude, hand pose, and posture.
We then fine-tune the generator, first trained on BEAT2, with AffectMoCap to retain broad conversational motion patterns while strengthening affective distinctions in generated body motion.

We evaluate SocialHumanoid at the levels of motion generation, affective perception, and physical execution.
On BEAT2, SocialHumanoid achieves competitive motion quality and speech-motion synchrony while substantially reducing inference latency through one-step generation.
Motion-only perceptual studies demonstrate more recognizable affective expression after AffectMoCap fine-tuning.
Real-robot experiments further verify continuous affect-conditioned behavior and stable long-horizon execution.

Our main contributions are summarized as follows:

1. We present \textbf{SocialHumanoid}, a real-time framework that connects affect-conditioned co-speech motion generation with continuous physical behavior on a humanoid robot and supports repeated response generation across dialogue turns.

2. We collect a professional affective co-speech motion-capture dataset, \textbf{AffectMoCap}, containing 4 hours of synchronized speech, full-body motion, fine-grained hand motion, and eight emotion annotations, providing high-quality supervision for learning distinguishable affective body expression.

3. We adapt improved MeanFlow to emotion-conditioned co-speech motion generation in a part-wise latent space, enabling continuous full-body synthesis through one-step inference per motion window while maintaining competitive motion quality and speech-motion synchrony.

4. Extensive evaluations demonstrate recognizable affective expression, low generation latency, and competitive motion quality, while real-robot experiments verify continuous affect-conditioned behavior and stable long-horizon execution.

\section{Related Work}

\subsection{Expressive Co-Speech Motion Generation}
Co-speech motion generation aims to synthesize body movements that are temporally and semantically aligned with speech.
Early learning-based methods directly mapped acoustic or multimodal speech features to upper-body gestures~\cite{yoon2020speech,li2021audio2gestures,liu2022learning}.
For co-speech motion specifically, subsequent work introduced diffusion models, motion matching, retrieval augmentation, and global-constraint formulations to capture the one-to-many relationship between speech and motion~\cite{zhu2023taming,ao2023gesturediffuclip,alexanderson2023listen,yang2023qpgesture,mughal2025retrieving,zhang2026mitigating}.
Recent methods increasingly model holistic motion involving the face, hands, body, and global movement~\cite{beat,yi2023generating,disco,chu2024corrtalk,emage,diffsheg,liu2024towards,xu2024mambatalk,wang2025mmgt,gesturelsm,zhang2025semtalk}.
These advances establish strong motion-quality and synchronization baselines, but most evaluate expressiveness on virtual humans rather than physical humanoid behavior.

Beyond co-speech gestures, general human-motion generation has explored text-conditioned completion, masked motion modeling, and motion-language foundation models~\cite{zeng2025progressive,guo2024momask,motiongpt2}.
These works provide useful representations and generative priors for human motion, but they do not directly address speech-synchronized full-body behavior or physical humanoid execution.
In parallel, audio-semantic talking-head generation studies speech-driven facial animation~\cite{liu2024audio}, which is complementary to our focus on full-body co-speech motion and robot embodiment.

Controllable generation provides a direct mechanism for expressive body language.
Existing methods condition motion on reference examples, speaker identity, semantic attributes, style, or emotion~\cite{ghorbani2023zeroeggs,ao2023gesturediffuclip,chen2024enabling,chhatre2024emotional,liu2025semges,wu2026speech,shen2025ted,semconflow,zhou2026exges,jin2026sentiavatar,fang2026coordspeaker,zhang2026personagesture}.
Producing visibly different emotion-conditioned motion requires supervision in which affect is expressed consistently through gesture amplitude, motion energy, posture, and hand articulation.
BEAT~\cite{beat} provides broad conversational motion with emotion labels, while ZeroEGGS~\cite{ghorbani2023zeroeggs} captures diverse gestural styles.
SocialHumanoid complements these resources with AffectMoCap, which focuses on professional-actor performances under explicit emotion conditions and provides synchronized speech, full-body SMPL-X motion, and detailed hands.

\subsection{Efficient Co-Speech Motion Synthesis}
Many high-quality co-speech models operate offline or rely on iterative sampling, assuming access to a complete speech segment before generating the corresponding motion.
This assumption is inconvenient for physical humanoids, where speech playback and motion execution consume a continuous stream of references.
Recent work has therefore investigated efficient and online synthesis.
GestureLSM~\cite{gesturelsm} uses latent shortcut modeling to reduce sampling cost.
MIBURI~\cite{miburi} adopts a causal autoregressive model with body-part-aware hierarchical tokens, LiveGesture~\cite{livegesture} targets zero-look-ahead arbitrary-length generation, and ProAct~\cite{proact} uses streaming flow matching with asynchronous high-level intention control.
These methods demonstrate the importance of low-latency generation and temporal continuity.
SocialHumanoid focuses on one-step affect-conditioned motion synthesis: a part-wise latent representation and improved MeanFlow objective enable one-step inference for each motion window, while history tokens connect consecutive windows.

\subsection{Executable Co-Speech Behavior on Physical Humanoids}
Motion generated in a human representation cannot be executed directly by a robot with a different kinematic tree, body proportion, joint range, and actuation capability.
Motion retargeting adapts human skeleton or SMPL-X~\cite{pavlakos2019expressive} trajectories to robot references by preserving selected positions, orientations, end-effector trajectories, or contact states.
Classical methods formulate this process as inverse kinematics or constrained optimization, while learning-based and contact-aware approaches improve cross-skeleton transfer and reduce artifacts such as foot sliding~\cite{yang2025omniretarget,pan2025spider}.
For humanoid robots, General Motion Retargeting (GMR)~\cite{araujo2025retargeting} formulates human-to-robot transfer as a multi-objective inverse-kinematics problem and produces embodiment-compatible reference trajectories.

Retargeted references must then be tracked while maintaining balance, robustness, and physical feasibility.
Physics-based imitation and whole-body tracking methods~\cite{luo2023perpetual,ji2024exbody2,luo2025sonic,yang2025omniretarget,ze2025twist2,zetwist,zhu2026clot}, including generalist controllers such as SONIC, have expanded the range of whole-body behaviors that can be executed physically.
They are nevertheless commonly evaluated with predefined or offline references rather than motion arriving continuously from a co-speech generator.

Recent humanoid co-speech systems begin to connect generation and deployment.
PhysDrift~\cite{physdrift} addresses the embodiment gap through robot-native speech-to-motion generation, while ProAct~\cite{proact} combines low-latency synthesis with proactive behavior control.
RoboGesture~\cite{wang2026robogesture} builds a robot-centric semantic gesture dataset and integrates semantic-acoustic alignment, streaming motion generation, and collision-aware control for real-time humanoid interaction.
SocialHumanoid studies a complementary path toward expressive humanoid behavior: professional affective motion supervision, one-step co-speech generation in a human representation, and continuous deployment through online retargeting and whole-body control.

\section{SocialHumanoid System}
\begin{figure*}[t]
\centering
\includegraphics[width=\textwidth]{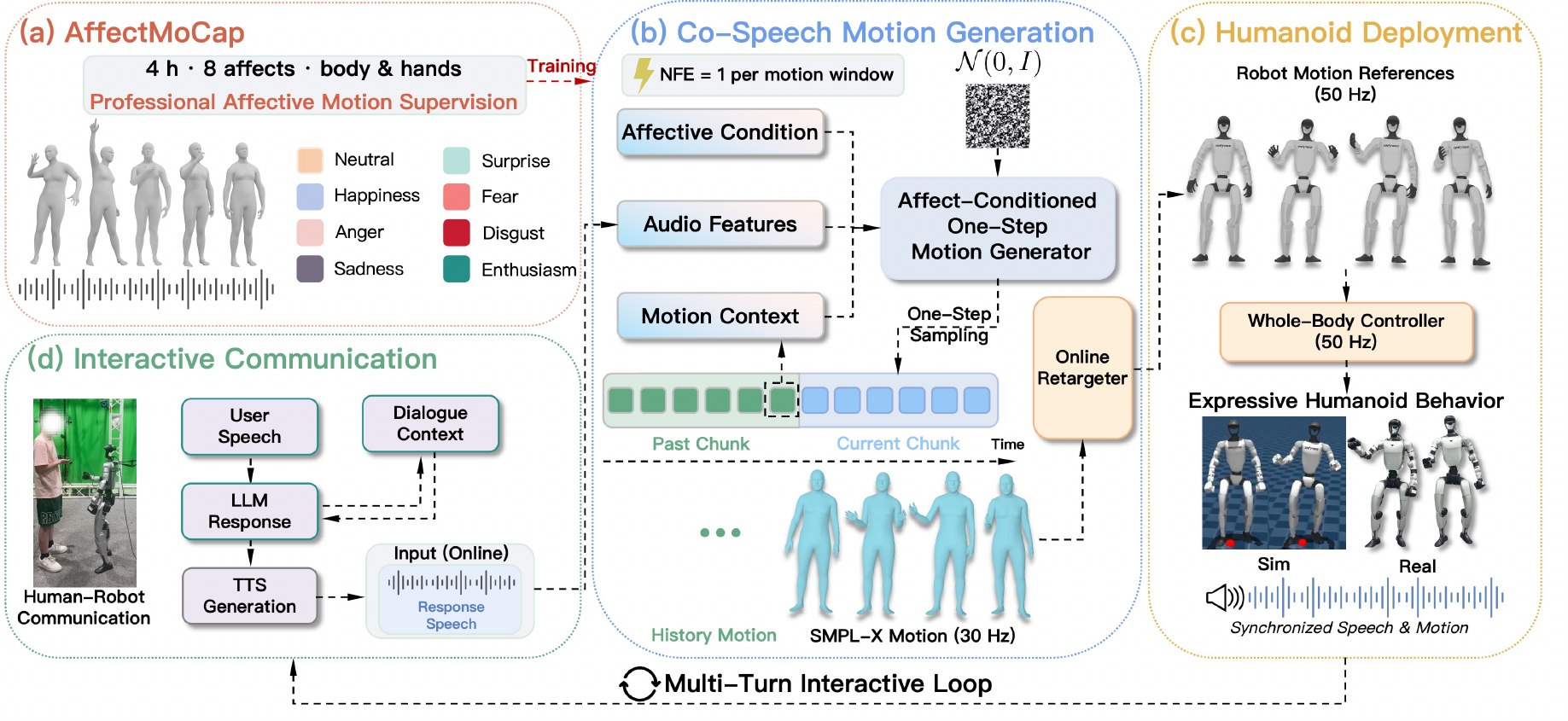}
\caption{\textbf{SocialHumanoid for expressive humanoid behavior.} (a) AffectMoCap provides professional affective full-body motion supervision for training. (b) During a robot response, audio features, an affective condition, Gaussian noise, and preceding motion context drive the affect-conditioned one-step generator, which produces each SMPL-X motion window with NFE $=1$ and links past and current chunks through history tokens. (c) Online retargeter and whole-body control convert the generated stream into synchronized behavior on the physical humanoid. (d) The interactive communication frontend maps user speech to an LLM response and TTS audio; after one response is executed, the system returns to the next user utterance and repeats the response pipeline.}
\label{fig:pipeline}
\end{figure*}

SocialHumanoid uses affect-conditioned one-step co-speech motion generation as the central path to expressive humanoid behavior.
Fig.~\ref{fig:pipeline} summarizes the system as four connected parts: affective supervision, co-speech motion generation, humanoid deployment, and interactive communication.
First, AffectMoCap provides 4 hours of professional full-body and hand-motion performances under eight affective conditions for supervising the motion generator.
During robot responses, audio features, an affective condition, Gaussian noise, and motion context from the preceding window drive the affect-conditioned one-step generator.
The generator synthesizes each full-body SMPL-X motion window through one-step inference, carries history tokens from past chunks to current chunks, and passes the generated stream to online retargeting.
The retargeted references are then tracked by a closed-loop whole-body controller on the physical humanoid.

At the communication level, user speech is processed by an LLM~\cite{yang2025qwen3,xu2026deepseek,singh2025openai} to produce a textual response, which TTS~\cite{hu2026qwen3} converts into response audio.
The response audio provides the online input for synchronized speech playback and co-speech motion generation during the robot response.
After the response is completed, the system returns to the listening state and applies the same response pipeline to the next user utterance, forming the turn-level interactive loop shown in Fig.~\ref{fig:pipeline} and supporting multi-turn dialogue.

\subsection{Affect-Conditioned One-Step Motion Generation}
The generator takes response speech and an affective condition as input and predicts a continuous sequence of full-body human motion.
Explicit emotion conditioning and AffectMoCap supervision encourage distinguishable body language, while the part-wise latent representation and improved MeanFlow objective allow each window to be generated in one forward process step.
Because the output remains in the human SMPL-X space, it is subsequently adapted to the target humanoid before physical execution.

\subsubsection{Part-Wise Latent Motion and Affective Conditioning}
Following~\cite{emage,gesturelsm}, we represent each human motion frame by a normalized SMPL-X motion vector, including 6D joint rotations, horizontal root-translation increments, and absolute root height, denoted as \(\mathbf{x}\in\mathbb{R}^{T\times 333}\).
Directly generating this high-dimensional motion in real time is difficult, so we first learn a compact part-wise latent motion space.
Specifically, the body is divided into upper-body, hands, and lower-body with translation groups.
For each group \(p\), a residual vector-quantized autoencoder~\cite{lee2022autoregressive} encodes the corresponding motion subset into a latent sequence,
\begin{equation}
    \mathbf{z}^{p}=E_{p}(\mathbf{x}^{p}), \quad
    \mathbf{z}=\mathrm{Concat}(\mathbf{z}^{upper},\mathbf{z}^{hands},\mathbf{z}^{lower})/s_{vq},
\end{equation}
where \(s_{vq}\) is a fixed latent scale.
The RVQ encoders and decoders are frozen when training the speech-driven generator.
This part-wise representation preserves detailed hand and upper-body motion while keeping the generation space small enough for online inference.

The condition encoder uses low-latency acoustic features extracted from the incoming waveform, including onset and amplitude cues.
A convolutional audio encoder maps the speech feature window to a latent resolution sequence.
To enable affective control, we add learnable emotion and speaker embeddings to the audio feature at each latent time step,
\begin{equation}
    \mathbf{c}_{i}=F_{a}(\mathbf{a})_{i}+\mathbf{e}_{emo(i)}+\mathbf{e}_{spk(i)} ,
\end{equation}
where \(\mathbf{a}\) is the audio feature, \(\mathbf{e}_{emo}\) is the affective embedding, and \(\mathbf{e}_{spk}\) models speaker-dependent gesture style.
Unknown label embeddings and conditional dropout are used during training so that the model remains robust when speaker or emotion information is missing.
For continuous generation, the first \(h\) latent tokens of each window are treated as a seed condition.
These seed tokens are copied from the previously generated window and are excluded from the loss, which enforces temporal continuity across adjacent chunks.

\subsubsection{Improved MeanFlow for One-Step Generation}
\begin{figure}[t]
\centering
\includegraphics[width=0.86\linewidth]{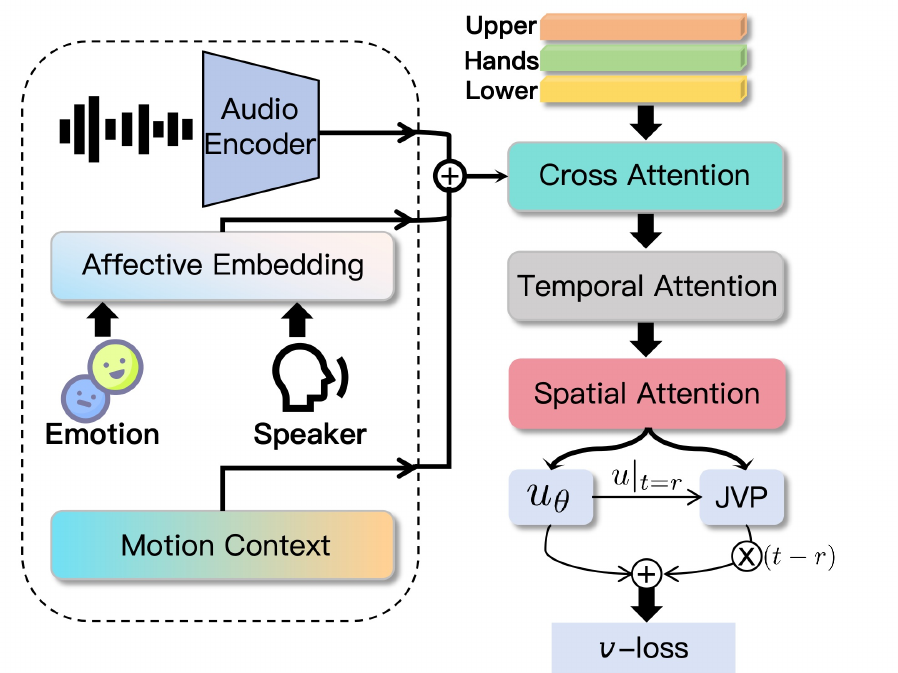}
\caption{\textbf{One-step co-speech motion generation.} The improved MeanFlow module predicts an interval velocity in the part-wise latent space, enabling one-step inference for each affect-conditioned motion window.}
\label{fig:improved_meanflow_sampling}

\end{figure}

We train the generator with an improved MeanFlow objective~\cite{geng2026improved} in the part-wise RVQ latent space.
Let \(\mathbf{z}_{0}\) be a clean motion latent and \(\boldsymbol{\epsilon}\sim\mathcal{N}(0,I)\) be Gaussian noise.
For a sampled time \(t\in[0,1]\), we construct a linear probability path
\begin{equation}
    \mathbf{z}_{t}=(1-t)\mathbf{z}_{0}+t\boldsymbol{\epsilon}, \quad
    \mathbf{v}_{t}=\boldsymbol{\epsilon}-\mathbf{z}_{0},
\end{equation}
where \(t=1\) corresponds to pure noise and \(t=0\) corresponds to the data latent.
Given \(\mathbf{z}_{t}\), the condition sequence \(\mathbf{c}\), the seed tokens, and an interval endpoint \(r\leq t\), a spatial-temporal Transformer denoiser predicts two velocity fields: an interval-averaged MeanFlow velocity \(\mathbf{u}_{\theta}\) and an auxiliary instantaneous velocity \(\mathbf{v}_{\theta}\).
The denoiser first processes the latent tokens separately for the three body groups, attends to the audio condition and affective condition through cross-attention, and then applies spatial-temporal Transformer blocks to model coordination among body parts and time.
As summarized in Fig.~\ref{fig:improved_meanflow_sampling}, the generator combines motion context, audio features, and affective embeddings before the cross-attention and spatial-temporal attention blocks, and the resulting dual-head outputs are optimized with the JVP-based improved MeanFlow objective.

The denoiser operates on part-wise RVQ latent streams for the upper body, hands, and lower body with root translation.
Let \(P\), \(L\), and \(d\) denote the numbers of body parts, latent positions, and latent channels, respectively.
Each RVQ token is projected to the latent channel dimension.
The seed tokens are flattened and projected; this seed embedding is added to embeddings of \(t\), the interval \(t-r\), and the training-time guidance tuple \((\omega,t_{\min},t_{\max})\), where \(\omega\) is the guidance scale and \([t_{\min},t_{\max}]\) is its active interval.
The resulting vector is concatenated with every motion token and fused by a linear layer.
Temporal rotary position embeddings (RoPE) are then applied independently to each body-part stream.
We denote the resulting tensor by
\(\mathbf{H}^{0}\in\mathbb{R}^{B\times P\times L\times d}\),
where \(B\) is the batch size.

\textit{Cross-attention.}
The part features are first concatenated along the channel dimension,
\(\mathbf{H}_{c}^{0}\in\mathbb{R}^{B\times L\times Pd}\).
The audio, emotion, and speaker condition sequence
\(\mathbf{C}\in\mathbb{R}^{B\times L\times Pd}\)
provides the keys and values, while the motion features provide the queries.
For attention head \(m\), the output is
\begin{equation}
    \mathbf{O}^{\mathrm{cross}}_{m}
    =
    \operatorname{Softmax}\!\left(
    \frac{\mathbf{Q}^{H}_{m}
    (\mathbf{K}^{C}_{m})^{\mathsf T}}{\sqrt{d_m}}
    \right)\mathbf{V}^{C}_{m}.
\end{equation}
Here, \(d_m\) denotes the feature dimension of each attention head.
The part dimension is restored after the cross-attention blocks.

\textit{Temporal attention.}
After reshaping the cross-attention output to
\(\mathbf{X}^{0}\in\mathbb{R}^{B\times P\times L\times d}\),
each spatial-temporal block first applies self-attention over the \(L\) latent positions independently for every body part.
After temporal RoPE, the output of head \(m\) for body part \(p\) is
\begin{equation}
    \mathbf{O}^{\mathrm{temp}}_{m,p}
    =
    \operatorname{Softmax}\!\left(
    \frac{\mathbf{Q}^{t}_{m,p}
    (\mathbf{K}^{t}_{m,p})^{\mathsf T}}{\sqrt{d_m}}
    \right)\mathbf{V}^{t}_{m,p}.
\end{equation}
This operation captures motion dynamics within each body-part latent stream.

\textit{Spatial attention.}
The temporal output is transposed so that attention is evaluated across body parts at each latent position.
After part-axis RoPE, the output of head \(m\) at latent position \(i\) is
\begin{equation}
    \mathbf{O}^{\mathrm{spa}}_{m,i}
    =
    \operatorname{Softmax}\!\left(
    \frac{\mathbf{Q}^{p}_{m,i}
    (\mathbf{K}^{p}_{m,i})^{\mathsf T}}{\sqrt{d_m}}
    \right)\mathbf{V}^{p}_{m,i}.
\end{equation}
Spatial attention exchanges information among body regions at each latent time step, while temporal attention preserves region-specific motion evolution.
The attention heads are concatenated and followed by an output projection, residual connection, pre-normalization, and MLP.
This temporal-then-spatial sequence is repeated across the spatial-temporal blocks.
Finally, separate output projections map the final features to \(\mathbf{u}_{\theta}\) and \(\mathbf{v}_{\theta}\).

Instead of only regressing the local flow velocity, we adopt the improved MeanFlow formulation and construct a compound velocity prediction using the total derivative of the average velocity along the guided target-velocity direction,
\begin{equation}
\begin{aligned}
    \mathbf{V}_{\theta}
    &=
    \mathbf{u}_{\theta}(\mathbf{z}_{t},t,r,\mathbf{c})
    +(t-r)\operatorname{sg}(\dot{\mathbf{u}}_{\theta}),\\
    \dot{\mathbf{u}}_{\theta}
    &=
    \left.
    \frac{d}{d\tau}
    \mathbf{u}_{\theta}
    (\mathbf{z}_{t}+\tau\mathbf{v}^{*},t+\tau,r,\mathbf{c})
    \right|_{\tau=0},
\end{aligned}
\end{equation}
and optimize
\begin{equation}
    \mathcal{L}_{gen}
    =
    \mathbb{E}_{\mathbf{z}_{0},\boldsymbol{\epsilon},t,r,\mathbf{c}}
    \left[
    \lambda_{u}\rho(\mathbf{V}_{\theta},\mathbf{v}^{*})
    +
    \lambda_{v}\rho(\mathbf{v}_{\theta},\mathbf{v}^{*})
    \right],
\end{equation}
where \(\dot{\mathbf{u}}_{\theta}\) is implemented with a Jacobian-vector product while keeping \(r\) and \(\mathbf{c}\) fixed, \(\operatorname{sg}(\cdot)\) stops the gradient through the JVP output, \(\rho(\cdot)\) is a masked robust loss, and \(\mathbf{v}^{*}\) is the guided target velocity.
The guidance target is obtained with classifier-free conditioning by comparing conditional and null-condition velocity predictions, which encourages the generated motion to follow speech rhythm while responding to the specified affective state.
The seed-token mask removes the copied history from the loss so that training focuses on the newly generated part of the window.

At inference time, the model starts from Gaussian latent noise \(\mathbf{z}_{t_s}=\boldsymbol{\epsilon}\) at the noise endpoint \(t_s=1\) and integrates the learned velocity field backward to the data endpoint \(t_e=0\).
Since the improved MeanFlow model learns an interval velocity, we perform one-step sampling in the real-time system,
\begin{equation}
    \hat{\mathbf{z}}_{0}
    =
    \mathbf{z}_{t_s}
    -
    (t_s-t_e)
    \mathbf{u}_{\theta}(\mathbf{z}_{t_s},t_s,t_e,\mathbf{c}),
\end{equation}
followed by clamping the history tokens to the seed latents.
For long speech sequences, the system repeatedly generates latent windows, appends only the non-history tokens, and uses the last \(h\) generated tokens as the seed for the next window.
Finally, the frozen part-wise RVQ decoders reconstruct upper-body, hand, and lower-body motion, and the decoded groups are merged into a full human motion stream for online retargeting.

\subsection{Online Human-to-Humanoid Motion Transfer}
The motion generator operates in a human SMPL-X space, while the physical humanoid has a different kinematic tree, body proportion, joint range, and control interface.
Each newly generated chunk is therefore converted into robot reference positions through online General Motion Retargeting (GMR)~\cite{araujo2025retargeting} and pushed to the execution queue.
We recover axis-angle SMPL-X poses from the generated 6D rotations, integrate the horizontal root-translation increments, directly recover the vertical root height, and use the SMPL-X body model to obtain per-frame body positions and orientations.
The motion is canonicalized by aligning its initial heading and horizontal origin before retargeting.

For each frame, GMR solves a constrained inverse-kinematics problem on the target robot model.
Let \(\mathbf{q}_{t}\) denote the robot configuration and \(\mathcal{M}\) the set of matched robot and human frames.
The objective minimizes their position and orientation errors:
\begin{equation}
    \mathbf{q}_{t}^{*}=\arg\min_{\mathbf{q}\in\mathcal{Q}}\sum_{m\in\mathcal{M}}\Big[w^{p}_{m}\|\mathbf{e}^{p}_{t,m}\|_{2}^{2}+w^{R}_{m}\left(e^{R}_{t,m}\right)^{2}\Big],
\end{equation}
where \(\mathcal{Q}\) is the feasible configuration set and \(\mathbf{e}^{p}_{t,m}\) and \(e^{R}_{t,m}\) are the mapped position and orientation errors.
We match the pelvis, feet, torso, shoulders, elbows, and wrists, emphasizing position for support-related frames and orientation for upper-body gesture frames.
Each solve is initialized from the previous configuration and constrained by the robot joint limits.

The resulting trajectory is ground-height adjusted, resampled to 50 Hz, and smoothed before entering the online motion queue.
Short interpolated transitions connect generated motion with the fallback pose at speech boundaries.
This queue allows generation and retargeting to prepare the next chunk while the controller executes the current reference stream.

\subsection{Physical Humanoid Behavior Execution}
Kinematically feasible references still require closed-loop stabilization on the physical robot.
We use a SONIC-based whole-body tracking controller~\cite{luo2025sonic} on a 29-DoF Unitree G1, treating the retargeted trajectory as a reference rather than a direct joint command.
Following its universal-token architecture, an encoder maps a 10-frame future window of joint positions, velocities, and anchor orientation to a command token.
The policy combines this token with a 10-frame proprioceptive history to predict the robot action:
\begin{equation}
    \mathbf{a}_{t}
    =
    \pi_{\phi}
    \left(
    \mathbf{o}_{t-H+1:t},
    \mathbf{g}_{t:t+K}
    \right),
\end{equation}
where \(\mathbf{o}_{t-H+1:t}\) and \(\mathbf{g}_{t:t+K}\) denote the 10-frame proprioceptive-history and future-command windows, respectively.
The policy follows the simulation tracking objective of SONIC, rewarding reference tracking while penalizing unstable or unsafe behavior.
At deployment, the 50 Hz policy consumes a sliding reference buffer, while a 500 Hz low-level thread publishes joint commands.
State freshness and joint velocity are monitored continuously, with damping commands used when execution stops.
Additional retargeting and control details are provided in the supplementary material.

\section{AffectMoCap: Learning Expressive Body Behavior}
\subsection{Overview}
AffectMoCap provides SocialHumanoid with synchronized speech and full-body motion under explicit affective conditions.
It is designed to supervise distinguishable co-speech body behavior rather than to maximize general-purpose motion scale alone.
The dataset is captured from two professional actors, one male actor and one female actor, to provide expressive performances with clear affective styles.
To reduce content bias across emotions, all speaking scripts are generated by GPT-5.5~\cite{singh2025openai} with a unified prompt template conditioned on predefined emotion labels.
The label set contains neutral, happiness, anger, sadness, surprise, fear, disgust, and enthusiasm.
For each recording, the actor performs the generated sentence according to the assigned affect label, producing paired speech, full-body motion, transcript, speaker identity, and emotion annotation.
The dataset focuses on body and hand gestures for humanoid execution and does not contain facial expression, gaze, or lip-motion data.
Tab.~\ref{tab:dataset_comparison} compares AffectMoCap with representative co-speech motion datasets.

\begin{table}[t]
\centering
\caption{\textbf{Data supervision for expressive co-speech body behavior.} We compare AffectMoCap with representative datasets in terms of body scope, affect/style annotation, and recording scale. PGT-Mesh denotes SMPL or SMPL-X motion predicted or fitted from video, whereas MC-Mesh denotes mesh motion converted from motion-capture data. For the Embody 3D dataset~\cite{mclean2025embody}, we report statistics only for the Dyadic Conversations section.}
\label{tab:dataset_comparison}
\footnotesize
\setlength{\tabcolsep}{3pt}
\renewcommand{\arraystretch}{1.05}
\begin{tabularx}{\linewidth}{@{}>{\raggedright\arraybackslash}p{0.32\linewidth}>{\raggedright\arraybackslash}X>{\centering\arraybackslash}p{0.15\linewidth}>{\centering\arraybackslash}p{0.10\linewidth}@{}}
\toprule
\textbf{Dataset} & Body Scope & Affect Label & Scale \\
\midrule
Trinity Mocap~\cite{ferstl2018investigating} & Full body (3D) & -- & 4 h \\
S2G-2D~\cite{ginosar2019learning} & Upper body (2D) & -- & 60 h \\
TED-2D~\cite{yoon2019robots} & Upper body (2D) & -- & 97 h \\
TWH Mocap~\cite{lee2019talking} & Full body (3D) & -- & 20 h \\
TED-3D~\cite{yoon2020speech} & Upper body (3D) & -- & 97 h \\
S2G-3D~\cite{habibie2021learning} & Upper body (3D) & -- & 38 h \\
TED-3D+~\cite{liu2022learning} & Upper body (3D) & -- & 33 h \\
BEAT~\cite{beat} & Full body (3D) & 8 emotions & 76 h \\
ZeroEGGS~\cite{ghorbani2023zeroeggs} & Full body (3D) & 19 styles & 4 h \\
TED-SMPL~\cite{lu2023co} & Upper body (PGT-Mesh) & -- & 30 h \\
S2G-SMPL~\cite{yi2023generating} & Upper body (PGT-Mesh) & -- & 27 h \\
BEAT2~\cite{emage} & Full body (MC-Mesh) & 8 emotions & 60 h \\
Streamer~\cite{yang2025gesturehydra} & Upper body (PGT-Mesh) & -- & 58 h \\
SuSuInterActs~\cite{jin2026sentiavatar} & Full body (3D) & -- & 37 h\\
Embody 3D~\cite{mclean2025embody} & Full body (PGT-Mesh) & -- & 59 h\\
\midrule
\rowcolor{oursblue}
\textbf{AffectMoCap (ours)} & Full body (MC-Mesh) & 8 emotions & 4 h \\
\bottomrule
\end{tabularx}
\end{table}

\begin{figure*}[t]
\centering
\includegraphics[width=\textwidth]{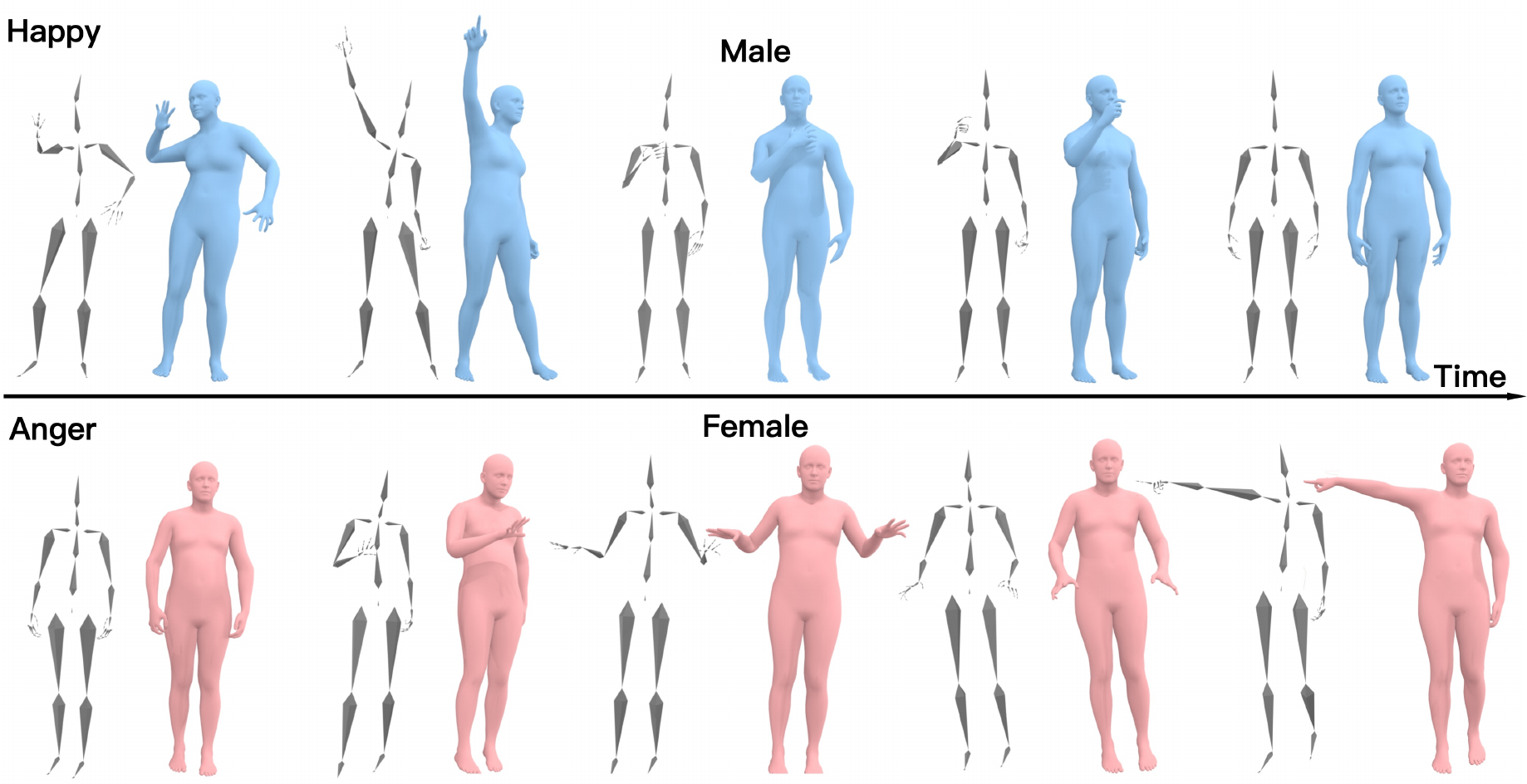}
\caption{\textbf{Representative samples from AffectMoCap.} The two rows show examples from the two professional actors. Each sample pairs the motion-capture skeleton with its fitted SMPL-X mesh, illustrating varied arm configurations, articulated hand poses, and full-body postures in the dataset.}
\label{fig:affectmocap_cases}
\vspace{-1.0em}
\end{figure*}

\begin{figure}[t]
\centering
\includegraphics[width=\linewidth]{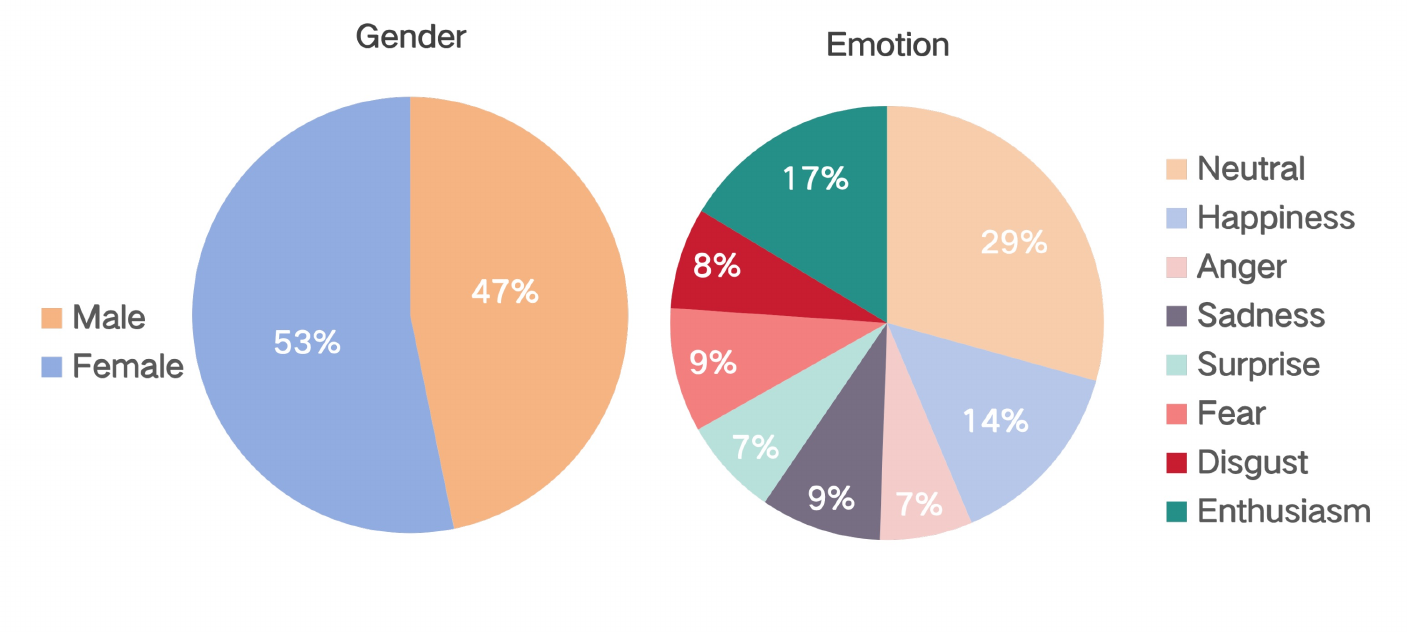}
\caption{\textbf{Duration distribution of AffectMoCap.} The left pie chart reports recording time by actor gender, and the right reports recording time across the eight emotion labels.}
\label{fig:dataset_duration_distribution}
\vspace{-1.0em}
\end{figure}

Beyond the aggregate statistics in Tab.~\ref{tab:dataset_comparison},
Fig.~\ref{fig:affectmocap_cases} presents representative samples from both actors, visualized as paired motion skeletons and fitted SMPL-X meshes.
Fig.~\ref{fig:dataset_duration_distribution} summarizes the recording-time distribution by actor gender and emotion category.

\subsection{Data Capture}
\begin{figure}[t]
\centering
\includegraphics[width=\linewidth]{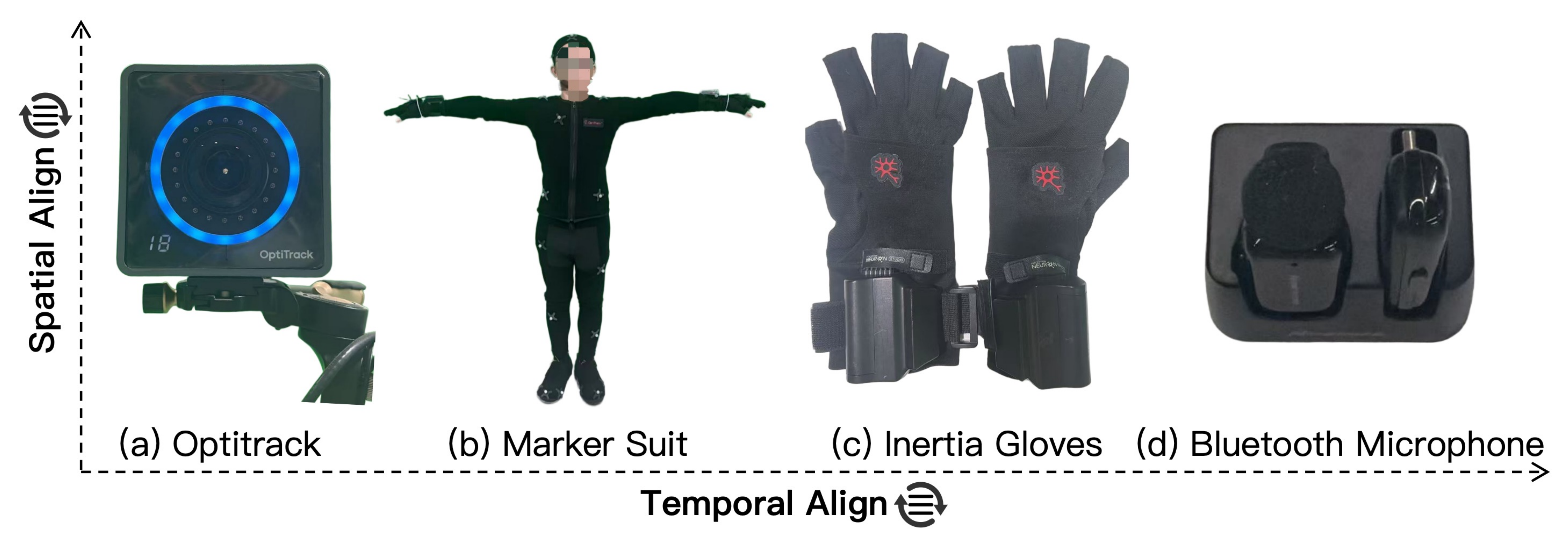}
\caption{\textbf{AffectMoCap capture setup.} The composite figure shows the hardware components used for synchronized affective co-speech capture: the OptiTrack optical tracking camera, the actor wearing a full-body marker suit, the hand motion-capture gloves, and the wireless microphone kit. All actors provided informed consent for data collection, analysis, and publication of anonymized results.}
\label{fig:affectmocap_setup}
\vspace{-1.0em}
\end{figure}

Motion is recorded in an OptiTrack~\cite{optitrack} motion-capture studio with an optical marker-based setup.
Before recording, each actor is calibrated to obtain a subject-specific skeleton and marker layout.
During capture, the actor wears an optical marker suit and hand motion-capture gloves~\cite{noitom}, stands in the capture volume, and performs the scripted sentences with the target emotion, while the system records full-body marker trajectories and synchronized speech.
Audio is captured with a wireless microphone at 48 kHz.
The recording protocol emphasizes upper-body expressiveness, hand articulation, posture, and body dynamics that are relevant to humanoid co-speech behavior.
Since the target system does not synthesize facial behavior, the capture setup does not record facial markers or facial video annotations.
The capture devices and actor configuration are illustrated in Fig.~\ref{fig:affectmocap_setup}.

After capture, the raw marker trajectories are cleaned by removing obvious marker swaps, filling short missing segments, and applying temporal smoothing to reduce optical jitter.
We then convert the cleaned marker motion into SMPL-X parameters.
For each actor \(s\), we estimate a subject-specific body shape \(\boldsymbol{\beta}_{s}\), and for each frame \(t\), we optimize the global translation \(\boldsymbol{\tau}_{t}\), global orientation, body pose, and hand pose so that the corresponding SMPL-X markers and joints align with the captured marker trajectories:
\begin{equation}
    \min_{\boldsymbol{\beta}_{s}, \boldsymbol{\Theta}_{t}}
    \mathcal{L}_{mk}
    +
    \lambda_{pose}\mathcal{L}_{pose}
    +
    \lambda_{temp}\mathcal{L}_{temp},
\end{equation}
where \(\boldsymbol{\Theta}_{t}\) denotes the per-frame SMPL-X pose and root parameters, \(\mathcal{L}_{mk}\) measures marker and joint alignment error, \(\mathcal{L}_{pose}\) regularizes implausible body and hand poses, and \(\mathcal{L}_{temp}\) encourages temporal smoothness.
Face-related SMPL-X parameters are fixed to a neutral state and are not used as supervision.

\subsection{Dataset Components}
Each processed sample contains a speech waveform, a full-body SMPL-X motion sequence, a transcript, an emotion label, and speaker metadata.
For compatibility with BEAT2-style co-speech motion data, the original 48 kHz audio is downsampled to 16 kHz, and each motion frame is converted to a 333-dimensional SMPL-X representation:
\begin{equation}
    \mathbf{x}_{t}
    =
    \left[
    \mathbf{R}^{6D}_{t,1},\ldots,\mathbf{R}^{6D}_{t,55},
    \Delta x_{t},h_{t},\Delta z_{t}
    \right],
    \quad
    \mathbf{x}_{t}\in\mathbb{R}^{333},
\end{equation}
where \(\mathbf{R}^{6D}_{t,j}\) denotes the 6D rotation representation of the \(j\)-th SMPL-X joint, \(\Delta x_t=\tau^x_t-\tau^x_{t-1}\) and \(\Delta z_t=\tau^z_t-\tau^z_{t-1}\) are frame-to-frame horizontal root displacements, and \(h_t=\tau^y_t\) is the absolute root height.
The first 330 dimensions encode 55 SMPL-X joint rotations in 6D form, and the last three dimensions follow the mixed root-translation representation \([\Delta x_t,h_t,\Delta z_t]\).
This format matches the input and output space of the co-speech motion generator, allowing AffectMoCap to be used together with the BEAT2 dataset without additional conversion.

\section{Experiments}
We evaluate whether one-step co-speech motion generation can support expressive humanoid behavior at two levels.
At the generation level, we measure motion quality, speech synchrony, one-step inference cost, emotion/style recognition, and the effect of AffectMoCap fine-tuning.
At the deployment level, we pass generated motions through a shared retargeting and whole-body control pipeline and evaluate affect-conditioned robot responses and runtime.
This separation distinguishes properties of the generated human-space motion from observations of the deployed humanoid behavior.

\subsection{Evaluation Protocol}
\subsubsection{Datasets and Baselines}
We first train and evaluate SocialHumanoid on the BEAT2 dataset~\cite{emage} and then fine-tune it on AffectMoCap to introduce clear affective supervision.
BEAT2~\cite{emage} provides 60 hours of high-quality SMPL-X-based gesture data from 25 speakers, including 12 female and 13 male speakers. It consists of 1,762 sequences depicting responses to everyday questions, with an average duration of 65.66 seconds per sequence. We incorporate emotion labels from BEAT~\cite{beat} into BEAT2.
We follow the common BEAT2 preprocessing protocol used by EMAGE~\cite{emage}: audio is represented at 16 kHz, motion is represented at 30 FPS, and each frame is converted to the SMPL-X representation described above.
The official train, validation, and test splits are used for quantitative evaluation so that our results are directly comparable with prior methods.

All AffectMoCap sequences are converted to the same 16 kHz audio and SMPL-X motion format as BEAT2.
To prevent the smaller affective dataset from being overwhelmed by BEAT2, AffectMoCap samples are drawn with balanced emotion sampling.
For evaluation, BEAT2 is used for standard objective metrics and baseline comparison, while held-out AffectMoCap clips are used for affective-controllability analysis and perceptual evaluation.

We compare SocialHumanoid with representative holistic co-speech motion generation baselines~\cite{liu2022learning,disco,beat,diffsheg,yi2023generating,liu2024towards,emage,xu2024mambatalk,chen2024enabling,gesturelsm, livegesture} reported on BEAT2.
For fair comparison, all methods are evaluated with the same BEAT2 Speaker-2 test split and the same motion representation when reproduced locally.
In addition to human-space gesture evaluation, we examine the deployed behavior by retargeting generated motions to the humanoid and executing them with the whole-body controller.

\subsubsection{Implementation Details}
Training proceeds in two stages on BEAT2~\cite{emage}, followed by an additional AffectMoCap fine-tuning stage for affective motion learning.
For each body region, the RVQVAE codebook is initialized uniformly with an embedding dimension of 128 and a codebook size of 1,024.
The RVQVAEs are trained for 30,000 iterations with a learning rate of \(2\times10^{-4}\).

The speech-to-motion generator contains 3 cross-attention layers and 8 spatial-temporal attention blocks.
Its latent dimension is 256, and its feed-forward dimension is 1024.
During speech-to-motion training, the RVQVAE codebooks remain frozen.
We train the generator for 1000 epochs on BEAT2 and then fine-tune it for 300 epochs on AffectMoCap.
We use the Adam optimizer with a learning rate of \(2\times10^{-4}\).
All model-training experiments are conducted on a single NVIDIA RTX 3090 GPU.
Additional generation and optimization hyperparameters are provided in the supplementary material.

\subsection{One-Step Co-Speech Motion Generation}
\textbf{Evaluation Metrics.}
We adopt the standard BEAT2 evaluation metrics used in recent co-speech gesture generation work: Fr\'echet Gesture Distance (FGD), Beat Consistency (BC), and Diversity.
These metrics evaluate motion realism, speech-motion synchrony, and gesture variability, respectively.
Since SocialHumanoid focuses on body motion for humanoid execution, we report body-motion metrics and runtime metrics as the core evaluation criteria.

\begin{table}[t]
\centering
\caption{\textbf{Motion quality and one-step sampling on BEAT2.} We report FGD \(\times 10^{-1}\), BC, Diversity, and the number of forward process steps (NFE) required to generate a 128-frame motion window.}
\label{tab:beat2_quantitative}
\setlength{\tabcolsep}{5pt}
\resizebox{\linewidth}{!}{
\begin{tabular}{lcccc}
\toprule
Methods & FGD (\(\downarrow\)) & BC (\(\uparrow\)) & Diversity (\(\uparrow\)) & NFE (\(\downarrow\)) \\
\midrule
Ground-Truth & -- & 0.703 & 11.97 & -- \\
\midrule
HA2G~\cite{liu2022learning} & 12.32 & 0.677 & 8.626 & 30 \\
DisCo~\cite{disco} & 9.417 & 0.643 & 9.912 & 1 \\
CaMN~\cite{beat} & 6.644 & 0.676 & 10.86 & 1 \\
DiffSHEG~\cite{diffsheg} & 7.141 & 0.743 & 8.21 & 25 \\
TalkShow~\cite{yi2023generating} & 6.209 & 0.695 & 13.47 & 64 \\
ProbTalk~\cite{liu2024towards} & 5.040 & 0.771 & 13.27 & 8 \\
EMAGE~\cite{emage} & 5.512 & 0.772 & 13.06 & 2 \\
MambaTalk~\cite{xu2024mambatalk} & 5.366 & 0.781 & 13.05 & 2 \\
SynTalker~\cite{chen2024enabling} & 4.687 & 0.736 & 12.43 & 1000 \\
GestureLSM~\cite{gesturelsm} & 4.088 & 0.714 & 13.24 & 8 \\
LiveGesture~\cite{livegesture} & 4.570 & 0.794 & 13.91 & 32 \\
\midrule
\rowcolor{oursblue}
\textbf{Ours} & \textbf{3.929} & 0.770 & 12.36 & \textbf{1} \\
\bottomrule
\end{tabular}
}
\end{table}

Because inference time depends strongly on the generated body scope and checkpoint configuration, we report AIST separately from the motion-quality metrics.
Tab.~\ref{tab:beat2_runtime} includes only methods for which compatible full-body checkpoints can be reproduced under the same evaluation protocol.
Methods providing only partial-body weights or runtime values measured with a different checkpoint configuration are excluded.

\begin{table}[t]
\centering
\caption{\textbf{Controlled generation-runtime comparison on BEAT2.} AIST is measured using the same 128-frame protocol on a single NVIDIA RTX 3090. Only methods with compatible full-body checkpoints are included.}
\label{tab:beat2_runtime}
\setlength{\tabcolsep}{10pt}
\begin{tabular}{lc}
\toprule
Methods & AIST (s) (\(\downarrow\)) \\
\midrule
DiffSHEG~\cite{diffsheg} & 4.4136 \\
TalkShow~\cite{yi2023generating} & 0.9279 \\
ProbTalk~\cite{liu2024towards} & 0.1472 \\
EMAGE~\cite{emage} & 0.1121 \\
MambaTalk~\cite{xu2024mambatalk} & 0.1580 \\
SynTalker~\cite{chen2024enabling} & 10.8981 \\
GestureLSM~\cite{gesturelsm} & 0.1734 \\
\midrule
\rowcolor{oursblue}
\textbf{Ours} & \textbf{0.0281} \\
\bottomrule
\end{tabular}
\end{table}

\textbf{Evaluation Results.}
As shown in Tab.~\ref{tab:beat2_quantitative}, SocialHumanoid achieves the best FGD among all compared methods, reducing FGD from 4.088 for GestureLSM to \textbf{3.929}.
This indicates a closer match to the real BEAT2 motion distribution.
Its BC of 0.770 is competitive with recent methods, although LiveGesture reports the highest value, and its Diversity of 12.36 remains close to the ground-truth value of 11.97.
SocialHumanoid therefore improves distributional realism while retaining motion variation and competitive rhythmic alignment.
These results establish the quality of the generated human-space motion; its contribution to expressive humanoid behavior is evaluated separately below through controlled perceptual and real-robot studies.
SocialHumanoid generates each 128-frame window in one forward process step, supporting low-latency delivery to the robot pipeline.
Under the controlled runtime protocol in Tab.~\ref{tab:beat2_runtime}, SocialHumanoid also achieves the lowest AIST among the reproducible full-body methods.

\subsection{Expressive Humanoid Behavior}
Automatic metrics characterize motion distribution, rhythmic alignment, and diversity in the human-motion space, but they do not establish whether an affect condition produces recognizable body language on a physical humanoid.
We therefore complement them with controlled emotion comparisons, robot-response visualizations, and two blinded perceptual evaluations.

\begin{figure*}[t]
\centering
\includegraphics[width=\textwidth]{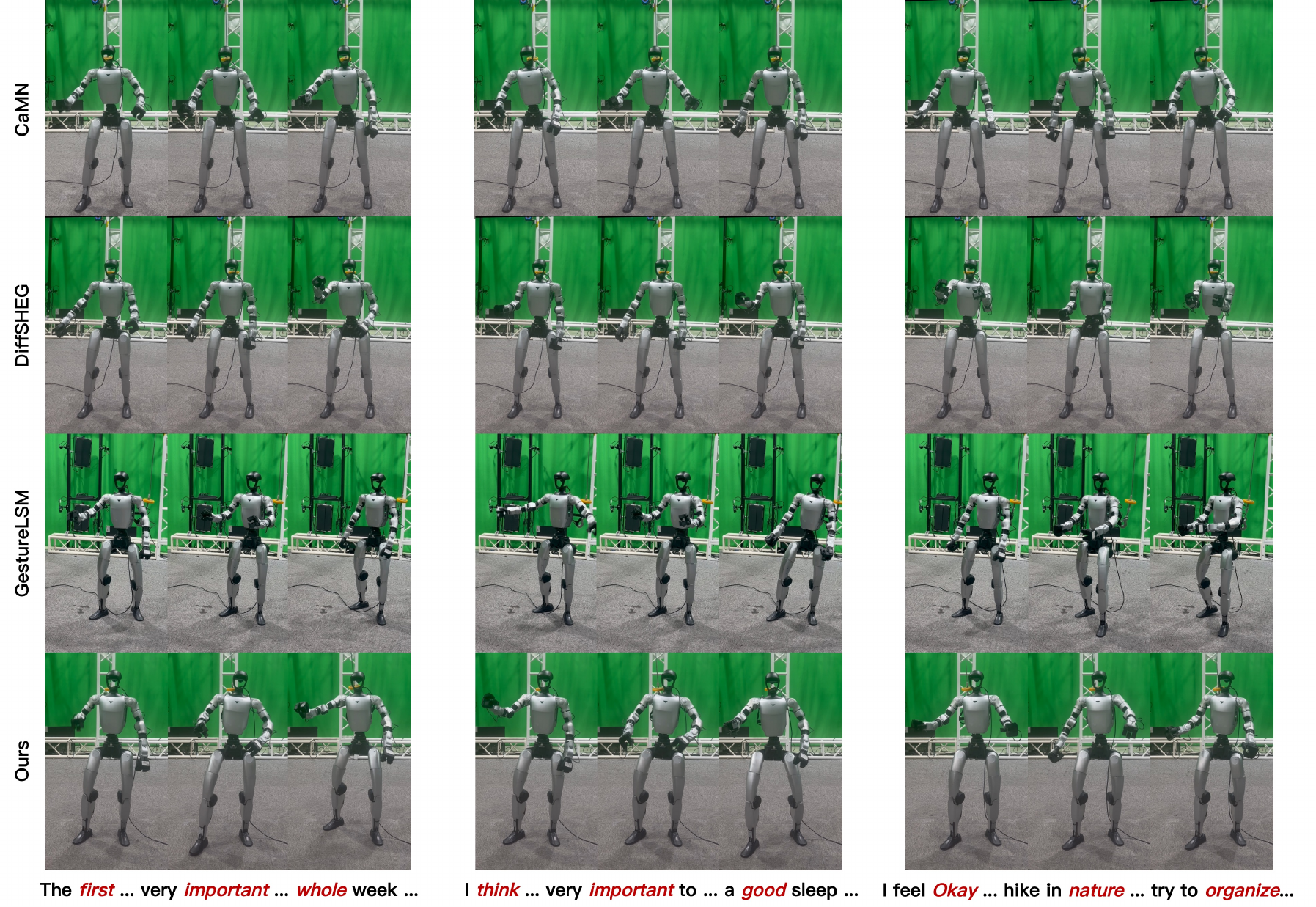}
\caption{\textbf{One-step co-speech motion deployed on the physical humanoid.} We compare methods using three BEAT2 test clips. Baseline motions are generated offline, whereas SocialHumanoid generates motion online; all outputs are evaluated through the same retargeting and whole-body control pipeline. Each column group shares the same speech clip, and red words indicate the corresponding speech content. Additional comparisons are provided on the project page.}
\label{fig:robot_results}
\end{figure*}

\begin{figure*}[t]
\centering
\includegraphics[width=\textwidth]{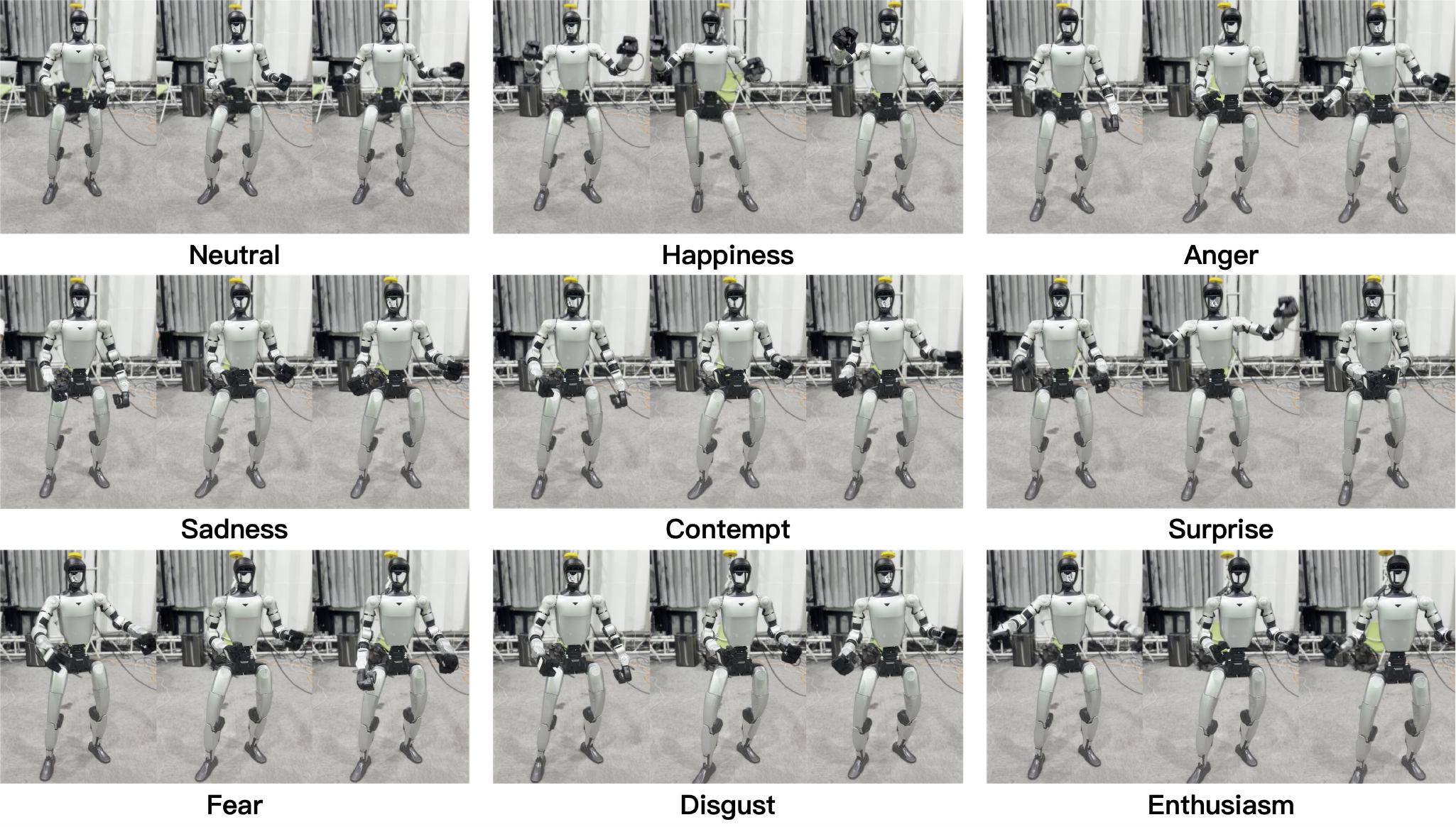}
\caption{\textbf{Affect-conditioned humanoid behavior under fixed speech input.} Rows correspond to different emotion labels, while speech audio, speaker identity, and displayed timestamps are fixed. Each column therefore compares robot poses at the same temporal position. Contempt is an additional affect label inherited from BEAT2 and is not included in the eight-label AffectMoCap annotation set.}
\label{fig:robot_affect_labels}
\end{figure*}

\textbf{Physical humanoid behavior.}
Fig.~\ref{fig:robot_results} compares CaMN, DiffSHEG, GestureLSM, and SocialHumanoid using three speech clips from the BEAT2 test set.
Because some baselines have insufficient inference efficiency for real-time generation, while others require TextGrid files produced by Montreal Forced Aligner (MFA), all baseline motion sequences are generated offline.
SocialHumanoid is the only method run online in this comparison.
For a controlled physical comparison, the resulting sequences from all methods are retargeted and tracked through the same robot execution pipeline described above, allowing direct comparison of their arm configurations and torso poses under identical speech content.
Fig.~\ref{fig:teaser} shows a turn-based response case in which the system converts a participant's request into a spoken robot response accompanied by continuous full-body motion.
The sequence demonstrates how dialogue generation and speech synthesis provide the response audio, while one-step motion generation, online retargeting, and whole-body control produce the corresponding physical body behavior.

\textbf{Emotion-conditioned motion.}
Fig.~\ref{fig:robot_affect_labels} isolates the effect of the emotion label by keeping the speech audio, speaker identity, and sampled timestamps unchanged.
In addition to the eight AffectMoCap categories, we visualize Contempt, an affect label inherited from the BEAT2 annotation space.
At corresponding timestamps, the conditions produce visibly different arm elevation, gesture openness, and torso posture.
This controlled comparison provides qualitative evidence of emotion-dependent body motion; the following perceptual evaluation tests whether viewers can recognize these differences without relying on speech cues.

\textbf{Emotion/style recognition from body motion.}
We conduct a blinded perceptual evaluation using real clips randomly sampled from BEAT2~\cite{emage}, ZeroEGGS~\cite{ghorbani2023zeroeggs}, and AffectMoCap, together with motion generated by SocialHumanoid.
The presentation order is randomized and source or model names are hidden.
Ten participants took part in the evaluation, with each participant viewing five randomly selected clips from each source or model.
For each clip, participants receive all emotion or style labels and select the best matching label. 
Each dataset and the full model are evaluated under paired audio+motion and motion-only conditions, allowing us to distinguish information conveyed by body motion from cues in the speech signal.
We additionally include a BEAT2-only generator under the motion-only condition to evaluate the effect of AffectMoCap fine-tuning, as summarized in Tab.~\ref{tab:user_study_affect_matching}.
All participants provided informed consent for data collection, analysis, and publication of anonymized results.

\begin{table*}[t]
\centering
\caption{\textbf{Perceptual evaluation of expressive co-speech motion.} Paired audio+motion and motion-only conditions separate speech cues from body-motion cues, while the BEAT2-only variant isolates the effect of AffectMoCap fine-tuning.}
\label{tab:user_study_affect_matching}
\setlength{\tabcolsep}{5pt}
\resizebox{\textwidth}{!}{
\begin{tabular}{lccc}
\toprule
Source  & Stimulus & Label Source  & Match Acc (\%) (\(\uparrow\))  \\
\midrule

BEAT2~\cite{emage}  & Motion only & BEAT emotion label  & 76.4  \\
BEAT2~\cite{emage}  & Audio + motion & BEAT emotion label  & 78.2  \\

ZeroEGGS~\cite{ghorbani2023zeroeggs}  & Motion only & Dataset style label  & 86.8 \\
ZeroEGGS~\cite{ghorbani2023zeroeggs}  & Audio + motion & Dataset style label  &  84.2 \\
AffectMoCap (ours)  & Motion only & Actor emotion label  & 88.6 \\

AffectMoCap (ours)  & Audio + motion & Actor emotion label  &92.4 \\
\midrule
SocialHumanoid (trained on BEAT2 only) & Gen. motion only & Target emotion label  & 80.2  \\

SocialHumanoid  & Gen. motion only & Target emotion label  & 82.7\\
SocialHumanoid  & Audio + gen. motion & Target emotion label  & 86.3 \\

\bottomrule
\end{tabular}
}
\end{table*}

Tab.~\ref{tab:user_study_affect_matching} shows that emotion or style remains recognizable from motion alone, with AffectMoCap achieving the strongest recognition among the real-motion datasets.
The full model also outperforms its BEAT2-only counterpart under the motion-only condition, while speech generally provides complementary cues.

\textbf{Robot response behavior.}
We separately evaluate SocialHumanoid during turn-based dialogue responses using fixed-view recordings of the physical humanoid.
Participants rate motion naturalness, agreement between the target emotion and the observed body language, and the perceived effect of system latency.
The deployed system is compared with variants that remove emotion conditioning, AffectMoCap fine-tuning, or speech playback, as well as the model retrained with MeanFlow.
Tab.~\ref{tab:user_study_robot_dialogue} reports five-point mean opinion scores; higher values are better for naturalness and emotion fit, whereas a lower delay-impact score indicates less disruption from latency.

\begin{table}[t]
\centering
\caption{\textbf{Perceptual evaluation of humanoid response behavior.} Naturalness and emotion fit are rated on a five-point scale, while delay impact measures perceived disruption from response latency on the same scale (lower is better).}
\label{tab:user_study_robot_dialogue}
\setlength{\tabcolsep}{5pt}
\resizebox{\linewidth}{!}{
\begin{tabular}{lccc}
\toprule
System setting & Naturalness (\(\uparrow\)) & Emotion Fit (\(\uparrow\)) & Delay Impact (\(\downarrow\)) \\
\midrule

w/o affective condition & 4.06 & 3.64 & 3.18 \\
w/o AffectMoCap fine-tuning & 3.94 & 3.30 & 3.22 \\
w/o speech playback & 4.12 & 3.62 & 3.56 \\
MeanFlow retraining and sampling & 4.10 & 3.98 & 3.28 \\
\rowcolor{oursblue}
SocialHumanoid & 4.23 & 3.92 & 3.14 \\
\bottomrule
\end{tabular}
}

\end{table}

Tab.~\ref{tab:user_study_robot_dialogue} shows that SocialHumanoid achieves the strongest naturalness and the lowest perceived delay impact, while AffectMoCap fine-tuning and affective conditioning improve emotion fit.
Although the MeanFlow variant receives a slightly stronger emotion-fit rating, SocialHumanoid provides better overall robot-response quality.

\textbf{Operational scope.}
Beyond spontaneous co-speech generation, we further test whether the deployed pipeline can support practical robot-response scenarios that are not fully captured by standard gesture benchmarks.
For explicit communicative behaviors such as greeting, pointing, counting, and directional gestures, we use a small retrieval library of 100 labeled action clips and execute the selected clips through the same retargeting and whole-body control pipeline.
This branch is used only for sparse predefined actions and is not treated as an output of the one-step generator.
We also test synthesized speech in English, Chinese, Japanese, and Spanish using the same audio-driven motion interface, showing that the pipeline can remain functional under different speech rhythms without language-specific motion modules.
Finally, a continuous speech sequence longer than 10 min is used to examine long-horizon execution.
The system repeatedly generates motion windows, carries history tokens across windows, and streams the retargeted references to the controller without interruption.
These cases are operational demonstrations rather than additional motion-generation benchmarks, with visual examples provided in the supplementary material.
Together, they indicate that SocialHumanoid can be embedded in a broader robot-response runtime while keeping the main technical contribution focused on affect-conditioned one-step co-speech motion generation.

\subsection{Component Analysis}
We analyze the components that support affect-conditioned body expression and one-step generation, together with the runtime cost of deploying the generated stream on the robot.
The model ablations evaluate the architecture, generation objective, and sampling strategy used by SocialHumanoid.
All ablated models are trained with the same data preprocessing and evaluated on the BEAT2 test split using FGD, BC, and Diversity.

\textbf{Effect of model architecture.}
As shown in Tab.~\ref{tab:ablation_components}, removing spatial attention causes the largest FGD increase, from 3.929 to 5.031, confirming the importance of coordination across upper-body, hand, and lower-body motion.
Removing temporal attention raises FGD to 4.648 despite a higher BC, indicating that local beat alignment alone does not ensure realistic temporal dynamics.
Without the affective condition, FGD increases to 4.730; its effect on perceived emotional control is evaluated more directly by the perceptual study in Tab.~\ref{tab:user_study_affect_matching}.
The full architecture achieves the best FGD while maintaining competitive BC and Diversity.

\textbf{Runtime of the deployed motion path.}
Tab.~\ref{tab:motion_module_runtime} reports stage-wise processing times recorded while the deployed motion pipeline runs on the physical robot with a single NVIDIA RTX 4090.
The reported values correspond to the three motion-related stages within the deployed pipeline: affect-conditioned motion generation, online retargeting from SMPL-X motion to humanoid joint targets, and whole-body controller execution.
Dialogue generation and speech synthesis remain active during the robot response but are excluded from the reported motion-module timings.

\begin{table}[t]
\centering
\caption{\textbf{Ablation study on BEAT2.} We report FGD \(\times 10^{-1}\), BC, Diversity, and NFE for model-architecture variants and retraining with MeanFlow or Shortcut.}
\label{tab:ablation_components}
\setlength{\tabcolsep}{4pt}
\resizebox{\linewidth}{!}{
\begin{tabular}{lcccc}
\toprule
Variant & FGD (\(\downarrow\)) & BC (\(\uparrow\)) & Diversity (\(\uparrow\)) & NFE (\(\downarrow\)) \\
\midrule
w/o temporal attention & 4.648 & 0.787 & 12.14 & 1 \\
w/o spatial attention & 5.031 & 0.783 & 12.51 & 1 \\
w/o affective condition & 4.730 & 0.768 & 12.88 & 1 \\
MeanFlow retraining and sampling & 4.078 & 0.769 & 13.89 & 1 \\
Shortcut retraining and sampling & 4.241 & 0.728 & 13.75 & 8 \\
\rowcolor{oursblue}
Full model & 3.929 & 0.770 & 12.36 & 1 \\
\bottomrule
\end{tabular}
}
\end{table}

\begin{table}[t]
\centering
\caption{\textbf{Runtime of the deployed motion path.} The processing time of each motion-related stage is measured during real-robot operation for a 128-frame generation window on a single NVIDIA RTX 4090.}
\label{tab:motion_module_runtime}
\setlength{\tabcolsep}{6pt}
\begin{tabular}{lc}
\toprule
Module & Time (s) \\
\midrule
Motion generation & 0.006 \\
Online retargeting & 1.248 \\
Whole-body controller computation & 0.008 \\
\bottomrule
\end{tabular}

\end{table}
\textbf{Effect of generation objective and sampling.}
Under the same preprocessing, MeanFlow retraining obtains an FGD of 4.078 versus 3.929 for improved MeanFlow; both retain 1-NFE sampling, but MeanFlow also produces a higher Diversity of 13.89.
Following the latent Shortcut formulation used by GestureLSM~\cite{gesturelsm}, Shortcut retraining obtains an FGD of 4.241 and a lower BC of 0.728 while requiring 8 NFE.
Although Shortcut produces higher Diversity, its weaker distributional quality, rhythmic alignment, and sampling efficiency favor the improved MeanFlow configuration for real-time deployment.

\section{Conclusion}
We presented SocialHumanoid, a system for expressive humanoid behavior via one-step co-speech motion generation.
The improved MeanFlow generator synthesizes each affect-conditioned full-body motion window in one forward process step, while history tokens maintain continuity across windows.
AffectMoCap provides professional full-body performances for learning visibly distinguishable affective expression, and online retargeting with whole-body control converts the generated SMPL-X stream into behavior on a physical humanoid.
Experiments on BEAT2 and AffectMoCap demonstrate competitive motion quality, speech--motion synchrony, recognizable affective body expression, and low generation latency.
Real-robot studies further show that the one-step motion stream can support continuous affect-conditioned responses and stable long-horizon execution after embodiment adaptation.

\bibliography{reference}
\bibliographystyle{IEEEtran}
\clearpage
\setcounter{section}{0}

\hyphenation{op-tical net-works semi-conduc-tor IEEE-Xplore}

\crefname{section}{Sect.}{Sects.}
\crefname{figure}{Fig.}{Figs.}
\crefname{table}{Tab.}{Tabs.}


\twocolumn[{%
    \centering

    {\normalfont\LARGE
    SocialHumanoid: Towards Expressive Humanoid Behavior
    via One-Step Co-Speech Motion Generation
    \par}

    \vspace{0.6em}

    {\normalfont\large
    Supplementary Material
    }

    \vspace{1.5em}
}]

\section{Improved MeanFlow Generation Details}
\label{sec:supp_imf_details}
We provide additional improved MeanFlow training details omitted from the main paper.

\subsection{Improved MeanFlow Training}
During training, the two interval endpoints are sampled from independent uniform variables and then ordered as
\begin{equation}
\begin{aligned}
    a,b &\sim \mathcal{U}(0,1),\\
    t = \max(a,b),&\quad
    r = \max(\min(a,b), t-\Delta_{\max}),
\end{aligned}
\end{equation}
where \(\Delta_{\max}\) caps the interval length.
To retain local flow-matching supervision, a fixed proportion \(p_{\mathrm{fm}}\) of each mini-batch is trained with \(r=t\).
For classifier-free guidance, the guidance scale and active guidance interval are also sampled during training:
\begin{equation}
\begin{aligned}
    \omega &= \exp(u\log(1+\omega_{\max})),\\
    t_{\min} &\sim \mathcal{U}(0,0.5),\quad
    t_{\max} \sim \mathcal{U}(0.5,1),
\end{aligned}
\end{equation}
where \(u\sim\mathcal{U}(0,1)\).
The effective scale used by the guided velocity target is
\begin{equation}
    \omega_{\mathrm{eff}}
    =
    \begin{cases}
    \omega, & t_{\min}\leq t\leq t_{\max},\\
    1, & \mathrm{otherwise}.
    \end{cases}
\end{equation}
The sampled scale and interval are embedded and passed to the denoiser, allowing the model to learn a range of guidance strengths instead of being tied to a single fixed value.

We use conditional dropout to make the model robust to missing speaker or affective information.
For dropped samples, the conditional input is replaced with the null condition and the guided target falls back to the unguided path velocity:
\begin{equation}
    \tilde{\mathbf{c}}=\emptyset,\quad \tilde{\mathbf{v}}^{*}=\mathbf{v}_{t}.
\end{equation}
The JVP output is detached before the corrected velocity loss is computed, while the auxiliary instantaneous-velocity head is trained with the same guided target.
This implementation stabilizes optimization while preserving the one-step inference path used in the real-time system.

\subsection{Training Convergence}
\begin{figure}[t]
\centering
\includegraphics[width=\linewidth]{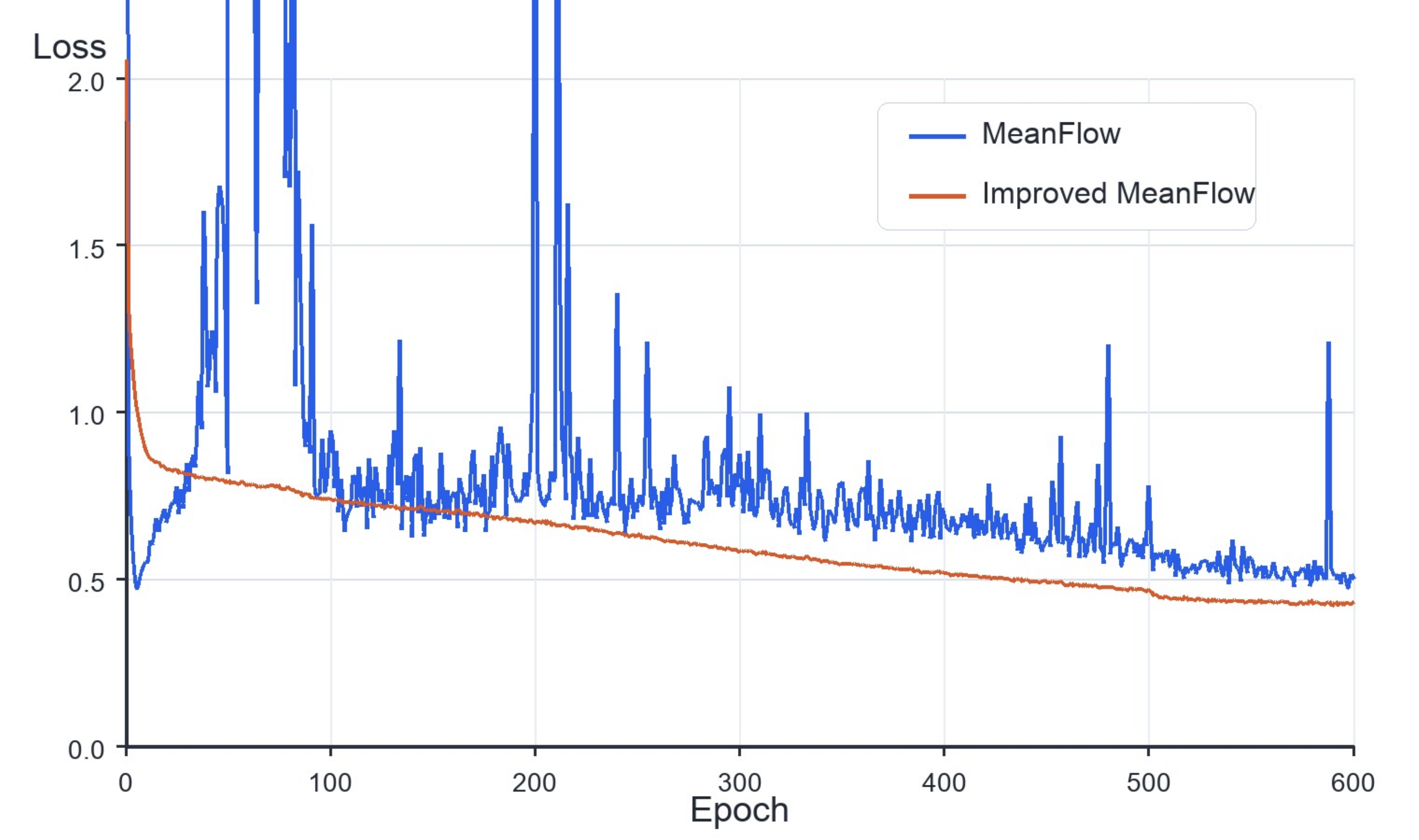}
\caption{\textbf{Training loss comparison.} MeanFlow and improved MeanFlow are compared over training epochs under the same training setup.}
\label{fig:supp_meanflow_loss_curve}
\end{figure}

Fig.~\ref{fig:supp_meanflow_loss_curve} compares the training losses of MeanFlow and improved MeanFlow under the same data preprocessing and training schedule.
Improved MeanFlow reaches the low-loss regime earlier and exhibits fewer large fluctuations, indicating faster and more stable optimization in our training setting.

\subsection{Generation Hyperparameters}
The remaining generation and optimization settings are summarized in Tab.~\ref{tab:supp_generation_hparams}.

\begin{table}[t]
\centering
\caption{\textbf{Key co-speech generation hyperparameters.} Values follow the improved MeanFlow training and streaming inference implementation.}
\label{tab:supp_generation_hparams}
\footnotesize
\setlength{\tabcolsep}{3pt}
\renewcommand{\arraystretch}{1.08}
\begin{tabular}{@{}>{\raggedright\arraybackslash}p{0.40\linewidth}>{\raggedright\arraybackslash}p{0.54\linewidth}@{}}
\toprule
\textbf{Hyperparameter} & \textbf{Value} \\
\midrule
History length \(h\) & 4 latent tokens (16 motion frames) \\
Motion window length & 128 frames (32 latent tokens) \\
Streaming hop length & 112 frames (28 latent tokens) \\
Training clip stride & 20 motion frames \\
RVQ temporal downsampling & \(4\times\) \\
Residual quantizer stages & 6 per body region \\
Maximum interval \(\Delta_{\max}\) & 1.0 \\
Training guidance cap \(\omega_{\max}\) & 1.0 (\(\omega\in[1.0,2.0]\)) \\
Loss weights \((\lambda_u,\lambda_v)\) & \((1.0,1.0)\) \\
Batch size & 64 \\
Transformer drop-path rate & 0.1 \\
Condition dropout rates & 0.1 \\
RVQ quantization dropout & 0.2 \\
CFG scale for evaluation & 1.8 \\
Audio features and rate & Onset and amplitude at 16 kHz; encoder-aligned to 30 Hz and pooled to 7.5 Hz \\
\bottomrule
\end{tabular}
\end{table}

\subsection{Classifier-Free Guidance Scale}
\begin{figure}[t]
\centering
\includegraphics[width=\linewidth]{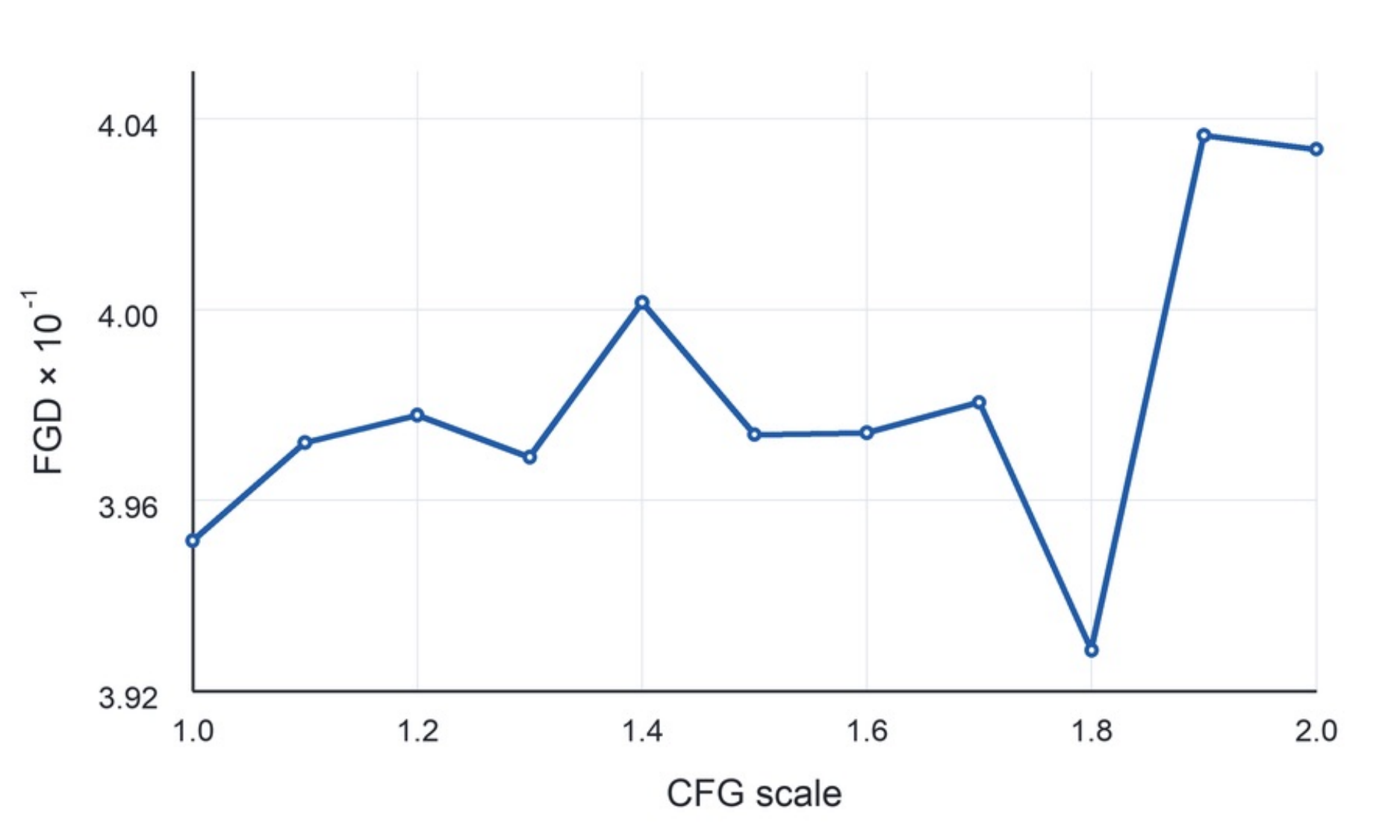}
\caption{\textbf{Effect of classifier-free guidance scale on FGD.} The scale is varied from 1.0 to 2.0 while all other inference settings remain fixed.}
\label{fig:supp_cfg_sweep_fgd}
\end{figure}

Fig.~\ref{fig:supp_cfg_sweep_fgd} evaluates guidance scales from 1.0 to 2.0 while all other inference settings remain fixed.
FGD varies non-monotonically across this range: it remains relatively stable from 1.0 to 1.7, reaches its lowest value at 1.8, and then increases at 1.9 and 2.0.
This trend indicates that stronger classifier-free guidance does not consistently improve distributional quality; an excessively large scale can instead move the generated motion away from the real-motion distribution.
Based on this sweep, we use a guidance scale of 1.8 in the main experiments.

\section{Retargeting and Control Implementation Details}
\label{sec:supp_robot_execution}
Generated 6D rotations are converted to axis-angle SMPL-X poses.
For the root representation \([\Delta x_t,h_t,\Delta z_t]\), the horizontal increments are integrated over time, while \(h_t\) is used directly as the absolute root height.
After heading and origin canonicalization, GMR is solved frame by frame with joint limits and initialization from the previous robot configuration.
The retargeted trajectory is shifted above the ground plane, resampled to 50 Hz using linear and spherical interpolation, and temporally smoothed before entering the execution queue.
Short interpolated segments connect generated motion with the fallback pose at speech boundaries.

Each queued frame contains the robot root pose and 29 actuated joint angles.
The deployment runtime estimates joint velocities, remaps the GMR joint order to the policy order, and maintains a sliding reference window; when excessive delay is detected, the window is reset to the latest chunk.
The token encoder and control policy are exported to ONNX and executed with TensorRT.
The policy action is interpreted as a scaled position offset from the default pose,
\begin{equation}
    q^{cmd}_{t,j}=q^{def}_{j}+s_{j}a_{t,\sigma(j)},
\end{equation}
where \(s_j\) is the joint-specific action scale and \(\sigma(\cdot)\) maps policy indices to the hardware joint order.
The 500 Hz low-level thread applies the target through a PD command,
\begin{equation}
    \tau_{t,j}=K^{p}_{j}(q^{cmd}_{t,j}-q_{t,j})-K^{d}_{j}\dot{q}_{t,j}.
\end{equation}
The runtime monitors state freshness and joint velocity limits and switches to damping commands when execution is stopped.

\section{Operational Scope}
\label{sec:supp_operational_scope}
Beyond spontaneous generated behavior, we examine predefined communicative actions, multilingual synthesized speech, and long-duration operation within the same retargeting and whole-body control pipeline.
These cases characterize the operational scope of SocialHumanoid and are not treated as additional motion-generation contributions.

\subsection{Predefined Communicative Actions}
\begin{figure}[t]
\centering
\includegraphics[width=0.9\linewidth]{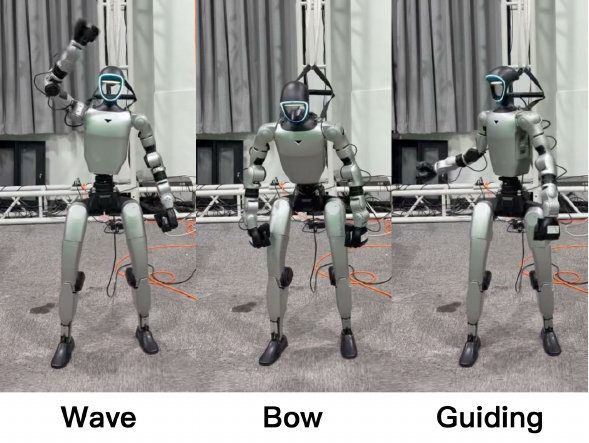}
\caption{\textbf{Predefined communicative robot actions.} Three representative action clips are executed on the physical humanoid. The clips use the same retargeting and control pipeline as generated co-speech motion and are not outputs of the one-step generator.}
\label{fig:supp_fine_grained_actions}
\end{figure}

General co-speech datasets provide limited supervision for explicit communicative actions such as greeting, pointing, counting, and directional gestures.
We therefore collect 100 short action clips with action labels and use them as a small retrieval library.
When a requested behavior matches an available label, the corresponding clip is inserted into the motion stream and executed through the standard robot pipeline.
This retrieval branch complements the generative model with actions that are sparsely represented in general co-speech data rather than treating them as learned outputs of the generator.
Representative real-robot cases are shown in Fig.~\ref{fig:supp_fine_grained_actions}.

\subsection{Multilingual Speech-Driven Execution}
\label{sec:supp_multilingual_execution}
\begin{figure*}[t]
\centering
\includegraphics[width=\textwidth]{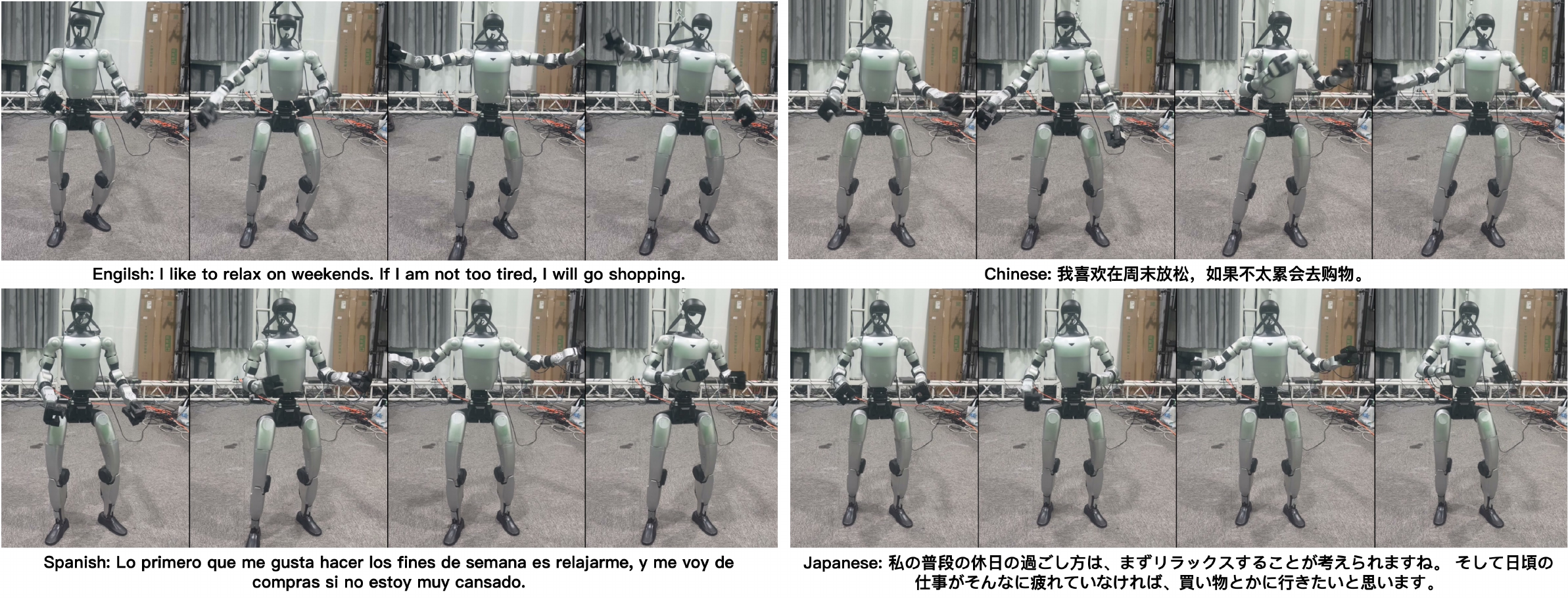}
\caption{\textbf{Multilingual speech-driven execution.} Real-robot sequences are driven by synthesized speech in English, Chinese, Japanese, and Spanish. Each block presents consecutive frames from one language condition using the same motion generation, retargeting, and control pipeline.}
\label{fig:supp_multilingual_results}
\end{figure*}

The audio-driven interface allows the deployed motion pipeline to process synthesized speech in English, Chinese, Japanese, and Spanish without introducing a language-specific motion module.
All four cases use the same motion generator, online retargeting module, and whole-body controller, without language-specific motion training or adaptation.
Because the generator uses acoustic onset and amplitude cues rather than text tokens, these trials primarily examine whether rhythmic motion generation and downstream execution remain functional across different spoken languages.
As shown in Fig.~\ref{fig:supp_multilingual_results}, each speech input produces a continuous motion sequence that is retargeted and executed on the physical humanoid.
These results are intended as qualitative demonstrations of system compatibility and do not constitute a quantitative benchmark of multilingual motion quality.

\subsection{Long-Duration Operation}
We evaluate long-duration operation using a continuous speech sequence longer than 10 min.
The generator repeatedly produces latent windows and carries the last \(h\) tokens of each window into the next, while retargeting and control consume the resulting motion stream online.
The system completes the extended sequence without interruption or loss of execution stability, showing that the window transition and motion queue support operation beyond short clips.

\section{Discussion and Limitations}
\label{sec:supp_discussion_limitations}
The system can repeat the response pipeline across dialogue turns, but the present study evaluates the body behavior displayed during robot responses rather than autonomous behavior planning, long-term interaction strategy, or adaptation to individual users.
AffectMoCap also remains limited in scale and label granularity, and the Unitree G1 lacks dexterous hands, so detailed SMPL-X finger gestures are reduced to arm motion, wrist orientation, and coarse end-effector movement.
Although motion generation uses one-step inference, online retargeting remains the dominant motion-processing cost during deployment.
Moreover, generation, retargeting, and control are modular rather than jointly optimized, so the human-space generator does not directly account for robot morphology or dynamics.
Future work will expand affective motion supervision, support dexterous hands, accelerate embodiment adaptation, and investigate tighter optimization between one-step generation and physical control.

\section{Evaluation Metrics}
\label{sec:supp_metrics}
Unless otherwise specified, the quantitative motion metrics are computed in the human SMPL-X motion space before retargeting, following the BEAT2 evaluation setting used in the main paper.

\subsection{Fr\'echet Gesture Distance}
Fr\'echet Gesture Distance (FGD) measures the distribution distance between generated and ground-truth motions in a learned gesture feature space.
Let \((\boldsymbol{\mu}_{g},\boldsymbol{\Sigma}_{g})\) and \((\boldsymbol{\mu}_{r},\boldsymbol{\Sigma}_{r})\) be the mean and covariance of generated and real gesture features, respectively.
FGD is computed as
\begin{equation}
    \mathrm{FGD}
    =
    \|\boldsymbol{\mu}_{g}-\boldsymbol{\mu}_{r}\|_{2}^{2}
    +
    \mathrm{Tr}
    \left(
    \boldsymbol{\Sigma}_{g}
    +
    \boldsymbol{\Sigma}_{r}
    -
    2(\boldsymbol{\Sigma}_{g}\boldsymbol{\Sigma}_{r})^{1/2}
    \right).
\end{equation}
Lower FGD indicates that the generated motions are closer to the real motion distribution and therefore more natural.

\subsection{Beat Consistency}
Beat Consistency (BC) evaluates temporal alignment between acoustic beats and motion beats.
Following the BEAT2 protocol, acoustic beats are detected from the speech onset envelope, and motion beats are detected from local minima of upper-body joint velocity.
For each acoustic beat \(b\), we find its nearest motion beat \(m\) and compute a Gaussian alignment score:
\begin{equation}
    \mathrm{BC}
    =
    \frac{1}{|\mathcal{B}_{a}|}
    \sum_{b\in\mathcal{B}_{a}}
    \exp
    \left(
    -\frac{\min_{m\in\mathcal{B}_{m}}\|b-m\|_{2}^{2}}
    {2\sigma^{2}}
    \right),
\end{equation}
where \(\mathcal{B}_{a}\) and \(\mathcal{B}_{m}\) denote the acoustic and motion beat sets, respectively.
Higher BC indicates better speech-motion rhythmic synchronization.

\subsection{Diversity}
Diversity measures the variability of generated gestures.
We compute the average pairwise distance between generated motion clips in the test set:
\begin{equation}
    \mathrm{Div}
    =
    \frac{2}{N(N-1)}
    \sum_{i<j}
    d(\mathbf{x}^{g}_{i},\mathbf{x}^{g}_{j}),
\end{equation}
where \(d(\cdot,\cdot)\) is the L1 distance in the motion feature space and \(N\) is the number of generated clips.
Following common BEAT2 reporting practice, Diversity should be interpreted together with FGD: a high Diversity score is desirable only when the generated distribution remains close to real motion.

\subsection{Runtime Metrics}
The number of function evaluations (NFE) reports the number of model evaluations used during inference; NFE $=1$ corresponds to one-step inference for a 128-frame motion window.
For runtime measurement, we report Average Inference Time per Sentence (AIST), computed on a single NVIDIA RTX 3090 GPU:
\begin{equation}
    \mathrm{AIST}
    =
    \frac{1}{S}
    \sum_{i=1}^{S}
    \left(t_{i}^{\mathrm{end}}-t_{i}^{\mathrm{start}}\right),
\end{equation}
where \(S\) is the number of evaluated sentences, and \(t_{i}^{\mathrm{start}}\) and \(t_{i}^{\mathrm{end}}\) denote the start and end timestamps of motion inference for the \(i\)-th sentence.
Lower NFE and AIST indicate better suitability for real-time streaming deployment.
For the runtime comparison, we include only methods with compatible public full-body checkpoints that can be evaluated on the same hardware and 128-frame protocol.
Methods with only partial-body weights or runtime measurements from a different checkpoint configuration are excluded rather than treated as directly comparable.




\end{document}